%% file: mlc_ranking_credal.tex
\documentclass[3p]{elsarticle}

\usepackage{soul}
\usepackage{lineno}
\usepackage{caption}
\usepackage{mathptmx}
\usepackage{helvet}
\usepackage{courier}
\usepackage{mathrsfs}
\usepackage{mathtools}
\usepackage{array}
\usepackage{blkarray}
\usepackage{amssymb}
\usepackage{amsmath,amsthm}
\usepackage{bbm}
\usepackage{comment}
\usepackage{tikz}
\usepackage{pgfplots}
\usepackage{bm}
\usepackage{algorithm}
\usepackage[algo2e]{algorithm2e}
\usepackage{hyperref}
\usepackage{float}
\usepackage{xcolor}
\usepackage{subfigure}
\usepackage{stmaryrd}
\usetikzlibrary{arrows.meta}

\newtheorem{theorem}{Theorem}
\newtheorem{proposition}[theorem]{Proposition}

\newtheorem{lemma}{Lemma}

\newtheorem{example}{Example}

\newtheorem{definition}{Definition}
\newtheorem{conjecture}{Conjecture}
\theoremstyle{plain}
\input{notations.tex}

\newcommand{\prob}{P}
\DeclareMathAlphabet{\mathpzc}{OT1}{pzc}{m}{it}

\newcommand{\spacelabel}{\mathscr{Y}}
\newcommand{\outspace}{\mathcal{K}}
\newcommand{\inputspace}{\mathscr{X}}
\newcommand{\data}{\mathscr{D}}
\newcommand{\outrank}{\mathscr{R}}
\newcommand{\loss}{\ell}
\newcommand{\lossrank}{\ell_R}
\newcommand{\vect}[1]{{\bm{#1}}}

\newcommand{\rank}[1]{\sum_{1\leq i < j \leq m}
		\mathbbm{1}_{(y_{#1(i)} = 0 \wedge y_{#1(j)} = 1)}}
\newcommand{\ranki}[3]{
	\mathbbm{1}_{(y_{#1(#2)} = 0 \wedge y_{#1(#3)} = 1)}
}

\newcommand{\tblue}[1]{\textcolor{blue}{#1}}
\newcommand{\newinstance}{\vect{x}}
\newcommand{\setn}[1]{\left[\vert #1 \vert\right]}
\newcommand{\setindices}{\mathscr{I}}

\newcommand{\credalind}{\mathfrak{P}}

\hypersetup{
    colorlinks=true,
    linkcolor=blue,
    filecolor=magenta,      
    urlcolor=cyan,
    citecolor=blue
}

\modulolinenumbers[5]
\journal{Journal of \LaTeX\ Templates}
\begin{document}
\begin{frontmatter}
\title{Skeptical inferences in multi-label ranking with sets of probabilities\tnoteref{t1}}

\tnotetext[t1]{This paper is a collaborative effort to investigate the Multi-label problem in an imprecise probabilistic setting. We will not be held responsible of any damages, injuries or losses that arise as a result of using our paper's results in part or in whole.}

\author{Yonatan Carlos {Carranza Alarc\'on}}
\ead{ycarranza.alarcon@gmail.com}

\author[tueaddress]{Vu-Linh {Nguyen}}
\ead{v.l.nguyen@tue.nl}

\address[tueaddress]{Department of Mathematics and Computer Science, Eindhoven University of Technology, The Netherlands}

\begin{abstract}
In this paper, we consider the problem of making skeptical inferences for the multi-label ranking problem. We assume that our uncertainty is described by a convex set of probabilities (i.e. a \textit{credal set}), defined over the set of labels. Instead of learning a singleton prediction (or, a completed ranking over the labels), we thus seek for skeptical inferences in terms of set-valued predictions consisting of completed rankings. 
\end{abstract}

\end{frontmatter}


\section{Introduction}

In contrast to multi-class classification problems where each instance is associated to one label, multi-label classification (MLC) consists in associating an instance to a subset of relevant labels from a set of possible labels. Such MLC problems arise in a number of problems including text categorization \citep{hayes1990construe,lewis1992evaluation}, music categorization \citep{trohidis2008multi}, semantic scene classification \citep{boutell2004learning}, or protein function classification \citep{elisseeff2002kernel}. We refer to \citep{zhang2013review} and \citep{tsoumakas2010random} for comprehensive survey articles on this~topic.

It is quite common in applications for the multi-label learner to output a \emph{ranking} on each query instance, that is, a ranking of labels from most likely relevant to most likely irrelevant. A prediction of that kind is commonly evaluated in terms of the rank loss which is the fraction of incorrectly ordered label pairs, where a relevant and a irrelevant label are incorrectly ordered if the former does not precede the latter \citep{dembczynski2012consistent,jung2018online,kanehira2016multi}.

The problem of making skeptical inferences for MLC under the presence of uncertainty has been studied in the literature~\cite{yonseb2020hammingmlc,nguyen2021multilabel,pillai2013multi}. \citet{pillai2013multi} seek to make skeptical inferences in terms of partial predictions, which are composed of a predicted part and an abstained part, where the abstained part basically captures the indices on those the multi-label learner is uncertain. To do that, \citet{pillai2013multi} focuses on maximizing the F-measure on the predicted part, subject to the constraint that the effort for manually providing the abstained part does not exceed a pre-defined value. The decision of whether or not to abstain on a label is guided by two thresholds on the predicted degree of relevance, which are learned via an empirical risk minimization principle. On the other hand, under the decision-theoretical perspective, \citet{nguyen2021multilabel} assume that the probabilistic predictions are made available, and seek the partial prediction by optimizing some generalized MLC metric, which is composed of the original metric on the predicted part and an additive penalty for the abstained part. In the case of rank loss, the partial ranking studied in \citep{nguyen2021multilabel} has a specific structure, in which its predicted part consists  of the highest-ranked and lowest-ranked labels. It is worth mentioning that the aforementioned works \citep{nguyen2021multilabel, pillai2013multi} assume a conventional probability setting, i.e., the estimation of a single distribution probability.

In contrast to the previous setting, we propose, in this paper, to study the problem of making skeptical inferences for the multi-label ranking problem from an \emph{imprecise} probabilistic approach. More precisely, we assume that our uncertainty is described by a convex set of probabilities, i.e. a \textit{credal set}~\citep{levi1980a}, defined over the set of labels. Instead of learning a singleton prediction, which is a completed ranking in the case of rank loss, we thus seek for skeptical inferences in terms of set-valued predictions, i.e., a set of completed rankings. In the imprecise probabilistic setting, there are different criteria~\citep{troffaes07decision} to define a set-valued predictions, including the well-known E-admissibility and maximality principles, which will be investigated in this paper.

Challenges in an imprecise probabilistic setting can arise either in the learning step, i.e. using an imprecise model learned from a training dataset, or in the inference step for efficiently finding the set-valued predictions which is our primary interest in this paper. Under the maximality principle, we prove that the expected difference of any pair of rankings can be expressed as a linear combination of the marginal probabilities whose weights are independent of the query instance. These theoretical results are then employed to propose approximate algorithms.  

In the next section, we briefly recall the setting of multi-label ranking and skeptical inferences with sets of probabilities. Section \ref{sec:theoretical_res} studies the problem of determining the set-valued predictions for the general case and a particular case of label independence where the joint conditional distribution factorises into the product of the conditional marginals. Experimental evaluation, as well as different discussions with the other approaches, will be a matter of future work.




\section{Preliminaries}
\label{sec:prel}
In this section, we introduce the necessary background to deal with our problem in a precise and imprecise probabilistic setting.

\subsection{Multi-label ranking}
In multi-label problem, given a set $\mathcal{K} =\{\lambda_1,\ldots,\lambda_m\}$, one assumes that to each instance $\newinstance$ of an input space $\mathcal{X}=\mathbf{R}^d$ is associated a subset $\Lambda_{\newinstance} \subseteq \Omega$ of relevant labels while its complement $\mathscr{K}\backslash\Lambda_{\newinstance}$ is considered as irrelevant ones for $\newinstance$. Let $\spacelabel = \{0, 1\}^m$  be a $m$-dimensional binary space and $\boldsymbol y = (y_1, \dots, y_m) \in \spacelabel$ be any element of $\spacelabel$ such that  $y_i = 1$ if and only if $\lambda_i \in \Lambda_x$, $0$ else.

Let us denote by $\succ$ the complete ranking (or an order relation) over the labels $\outspace$ (over $\outspace\times\outspace$, resp.). We can identify such ranking $\succ$ with a permutation function $\sigma:\setn{m}\rightarrow\setn{m}$\footnote{$\setn{m}=\{1, \dots, m\}$ is a set of the first $m$ integers.} such that
\begin{align}
	\sigma^{-1}(i) < \sigma^{-1}(j) \iff \lambda_i \succ_\sigma \lambda_j
\end{align}
where $\sigma(k)$ is the index $j$ of the label $\lambda_j$ and $k=\sigma^{-1}(j)$ is the position of label $\lambda_j$, besides $\lambda_i \succ_\sigma \lambda_j$ can be interpreted as $\lambda_i$ is preferred to $\lambda_j$ according to $\sigma$. We here consider that the labels $\lambda_*$ are ordered decreasingly, i.e. from the most to the least relevant (in other words, from the lowest to the highest position). Hence, its output space is the set $\outrank$ of complete rankings over $\mathcal{K}$ which contains $|\outrank|=m!$ elements (i.e., the set of all permutations). In this paper, we are interested in such kind of structured outputs, i.e. $\lambda_{i_1}\succ_\sigma\dots\succ_\sigma \lambda_{i_m}$, instead of an unordered vector of relevant labels, as is commonly investigated in the multi-label classification~\cite{zhang2013review,Read_2021,nguyen2021multilabel,yonseb2020hammingmlc}.

We assume that observations $\data = \{(\newinstance_i, \vect{y}_i)|i=1, \dots, N\}\subseteq\inputspace\times\mathcal{Y}$ are drawn i.i.d. from an unknown theoretical probability distribution $\prob:\mathcal{X} \times \mathcal{Y} \to [0,1]$, and denote $P_\newinstance(\bm{y}):= P(\bm{y}|\newinstance)$ the conditional probability of $\bm{y}$ given $\newinstance$. When our uncertainty is described by a (precise) estimated probability $\hat{\prob}$, obtained from fitting the set of observations $\data$ in a learning process, the goal of multi-label classification (as well as any other classification problem) with structured output space $\outrank$ is to pick the (prediction) ranking $\sigma'\in\outrank$ which minimizes the risk of getting missclassifications w.r.t. a specified instance-wise loss function $\loss:\outrank \times \mathcal{Y} \to \reals $ (cf. \cite[eq. 3]{dembczynski2012label} and \cite[eq. 2.21]{friedman2001elements}), where $\ell(\sigma',\vect{y})$ is the loss incurred by predicting $\sigma'$ when $\vect{y}$ is the ground-truth, i.e. 
\begin{equation}\label{eq:riskclassic}
\hat{\vect{y}}^{\hat{P}}_\ell=\underset{\sigma \in \outrank}{\arg\min~}\expe_{\hat{P}} \left(\ell(\sigma,\cdot) \right)=\underset{\sigma \in \outrank}{\arg\min} \sum_{\vect{y} \in \mathcal{Y}} \hat{P}_\insta(\vect{y}) \ell(\sigma,\vect{y})
\end{equation}
or, equivalently, by picking the maximal elements of the ordering $\sqsupseteq_{\ell}^{\hat{\prob}}$ where $\sigma_2 \sqsupseteq_{\ell}^{\hat{\prob}} \sigma_1$  ($\sigma_2$ is preferred to $\sigma_1$) if
\begin{align}\label{eq:compaprec} 
\expe_{\hat{\prob}} \left(\ell(\sigma_1,\cdot) - \ell(\sigma_2,\cdot) \right) &=  \sum_{\vect{y} \in \mathcal{Y}} \hat{\prob}_\insta(\vect{y}) \left( \ell(\sigma_1,\vect{y}) - \ell(\sigma_2,\vect{y}) \right) \nonumber\\
&=\expe_{\hat{P}} \left( \ell(\sigma_1,\cdot)\right) - \expe_{\hat{P}}\left(\ell(\sigma_2,\cdot)\right)\geq 0.
\end{align}
This equation means that exchanging $\sigma_1$ for $\sigma_2$ would incur a positive expected loss, due to the fact that expected loss of $\sigma_1$ is higher than $\sigma_2$, therefore $\sigma_2$ should be preferred to $\sigma_1$. Furthermore, since $\sqsupseteq_{\ell}^{\hat{\prob}}$ is a complete pre-order, picking any of the possibly indifferent maximal elements will be equivalent w.r.t. expected loss minimisation. Therefore, finding the maximal element(s) (or ranking(s)) will require $|\outrank|$ computations in general.

Given two rankings $\sigma_1$ and $\sigma_2$, we will denote by $\setindices_{\sigma_1\neq\sigma_2}:=\left\{(i,j) \middle| \lambda_i \succ_{\sigma_1} \lambda_j \text{~and~} \lambda_j \succ_{\sigma_2} \lambda_i \right\}$ the set of pairwise indices over which the rankings $\sigma_1$ and $\sigma_2$ have some disagreements. By disagreements, we mean those \emph{pairwise label preferences} of which the ranking $\sigma_1$ (or $\sigma_2$) has an \emph{opposite one} in the ranking $\sigma_2$ (resp. $\sigma_1$) (i.e., the preference of labels $\lambda_i\succ_{*}\lambda_j$ are swapped $\lambda_j\succ_{*}\lambda_i$.), regardless of how many other labels are between them (e.g. $\lambda_i\succ_{\sigma_1}\dots\succ_{\sigma_1}\lambda_j$ and $\lambda_j\succ_{\sigma_2}\dots\succ_{\sigma_2}\lambda_i$). Consequently, the set denoted by  $\setindices_{\sigma_1=\sigma_2}=\left\{(i,j) \middle| \lambda_i \succ_{\sigma_1} \lambda_j \text{~and~} \lambda_i \succ_{\sigma_2} \lambda_j \right\}$ captures those \emph{pairwise labels preferences} of which the ranking $\sigma_1$ and $\sigma_2$ agree on. Let us illustrate this matter in the next example. 
\begin{example}\label{exa:setdisagreements}
Let us consider a set of labels $\mathcal{K}=\{\lambda_1, \dots, \lambda_4\}$ and two rankings $\sigma_1$ and $\sigma_2$ over $\mathcal{K}$ defined as follows 
\begin{align*}
	&\lambda_1\succ_{\sigma_1}
	\lambda_3\succ_{\sigma_1}
	\lambda_4\succ_{\sigma_1}
	\lambda_2,\\
	&\lambda_2\succ_{\sigma_2}
	\lambda_1\succ_{\sigma_2} 
	\lambda_3\succ_{\sigma_2}  
	\lambda_4.
\end{align*}
The set of disagreements (or the set of pairwise indices) between $\sigma_1$ and $\sigma_2$ is
\begin{align*}
	\setindices_{\sigma_1\neq\sigma_2}=\{(1,2), (3, 2), (4, 2)\}, 
\end{align*} 
and the set of agreements is 
\begin{align*}
	\setindices_{\sigma_1=\sigma_2}=\{(1,3), (1, 4), (3, 4)\}, 
\end{align*} 
\end{example}

As in this paper, our goal is to make a set of predictions over all possible complete rankings $\outrank$, we will use the notation $\mathbb{R}_{\ell,\credal}\in\outrank$ to represent it. In what follows, we will see a way of how we can obtain such set-valued predictions by using set of probabilities.


\subsection{Skeptic inferences with distribution sets}
\subsubsection{Uncertainty representation} 
In this paper, we assume that our uncertainty is described by a convex set of probabilities $\credal$, a \textit{credal set}~\cite{levi1980a}, defined over $\mathcal{Y}$. Such sets can arise in different ways, either as a native result of the learning method~\cite{augustin2014introduction}, as the result of an agnostic\footnote{With respect to the missingness process.} estimation in presence of imprecise data, or as a neighbourhood taken over an initial estimated distribution $\hat{\prob}$~\cite{montes2020unifying,rahimian2019distributionally}. Given such a set of probabilities, we can define for any event $A \subseteq \mathcal{Y}$ the notions of lower and upper probabilities $\lpr(A)$ and $\upr(A)$, respectively as
\begin{equation*}
\lpr(A)=\inf_{\prob \in \credal_\insta} P(A) \quad\textrm{ and }\quad \upr(A)=\sup_{\prob \in \credal_\insta} P(A)	
\end{equation*}
with $P$ being precise probability measures. Lower and upper probabilities are dual, in the sense that $\lpr(A)=1-\upr(A^c)$. Similarly, if we consider a real-valued bounded function $f:\mathcal{Y}\to \reals$, the lower and upper expectations $\lexpe(f)$ and $\uexpe(f)$ are defined as
\begin{equation*}
\lexpe(f)=\inf_{P \in \credal} \expe(f) \quad\textrm{ and }\quad \uexpe(f)=\sup_{P \in \credal} \expe(f)	
\end{equation*}
where $\expe(f)$ is the precise expectation of $f$ w.r.t. $P$. 

\subsubsection{Skeptic inference and decision} 

Once our uncertainty is described by a credal set $\credal$, instead of a single probability $\prob$, the decision rule of Equation~\eqref{eq:compaprec} is no longer directly applicable, and therefore, it is necessary to use an extended version that benefits from strong theoretical justifications~\cite{troffaes07decision}. In this paper, we will focus on two different extensions which may return more than one solution in case of high uncertainty: E-admissibility and Maximality. 

\begin{definition}\label{def:E-adm} E-admissibility returns the set of predictions that are optimal for at least one probability within the set $\credal$. In other words, the E-admissibility rule returns the prediction set
\begin{equation}\label{eq:E-adm-set}
\hat{\mathbb{R}}^E_{\ell,\credal}=\condset{\sigma \in \mathcal{Y}}{\exists P \in \credal \textrm{ s.t. } \sigma=\hat{\sigma}^{P}_\ell}.
\end{equation}
\end{definition}

\begin{definition}\label{def:maxi} Maximality consists in returning the maximal, non-dominated elements of the partial order $ \sqsupset_{\ell,\credal}$ such that $\sigma \sqsupset_{\ell,\credal} \sigma'$ if
\begin{equation}\label{eq:compaIP}
\inf_{P \in \credal} \expe_{P} \left(\ell(\sigma',\cdot) - \ell(\sigma,\cdot) \right) > 0,
\end{equation}
that is if exchanging  $\sigma'$ for $\sigma$ is guaranteed to give a positive expected loss. The maximality rule returns the prediction set
\begin{equation}\label{eq:maximalset}
\hat{\mathbb{R}}^{M}_{\ell,\credal}=\condset{\sigma \in \outrank}{\not \exists \sigma' \in \outrank \textrm{ s.t. } \sigma' \sqsupset_{\ell,\credal} \sigma }.
\end{equation}
\end{definition}

These decision rules follow a skeptical strategy, in the sense that the set of solutions that they return is guaranteed to contain the optimal prediction, whatever the true distribution within $\credal$. Moreover, \citet{troffaes07decision} showed the set of solution given by E-admissibility is a subset of the one given by the Maximality, i.e. $\hat{\mathbb{R}}^E_{\ell,\credal} \subseteq \hat{\mathbb{R}}^M_{\ell,\credal}$.


Note that the maximality rule is well-known for being a more conservative skeptical decision rule than the E-admissibility rule. Yet, in terms of computational complexity, the E-admissibility set  $\hat{\mathbb{R}}^E_{\ell,\credal}$ is harder to compute than the maximality set $\hat{\mathbb{R}}^{M}_{\ell,\credal}$~\cite[\S 8]{augustin2014introduction}. Consequently, making predictions with probability sets is often harder than with precise ones, as it needs to solve complex optimization problems in the learning and inference steps. Hence, naively verifying each possible pair of rankings $\sigma',\sigma\in\outrank$ for the maximality rule is not practically possible\footnote{Similarly, finding the solution of the E-admissibility rule is impractical, since it may be done by naively enumerating the elements of $\outrank$ for each probability distribution $P$ in the set~$\credal$ in order to obtain the maximal \emph{ranking} element $\hat{\sigma}$ w.r.t ~$P$.} (i.e., a complexity of $\mathcal{O}(|\outrank|^2)$), and in the next section, we will show new and improved procedures for the general sets $\credal$ and then for specific sets (or constrained credal sets) induced from binary relevance models. 

\section{Skeptic inference for the Ranking loss}
\label{sec:theoretical_res}
The ranking loss is a quite common function used for comparing structured objects (e.g. a ranking $\sigma$ and a structured observed output $\vect{y}$) that do not have the possibility of having ties. By comparison, in the context de multi-label problem, we mean that it counts the number of pairs of labels that disagree between the ranking $\sigma$ and the partial order induced by $\vect{y}$ (assuming that all relevant labels are preferred to non-relevant ones), and it can be written as follows:
\begin{equation}
	\ell_R(\sigma, \vect{y}) =
	\sum_{\substack{(i,j)\in[m]\times[m]\\ y_i > y_j}}
		\mathbbm{1}_{\lambda_i \prec_\sigma \lambda_j},
\end{equation}

where $\indicator{A}$ denotes the indicator function of event $A$ and $\lambda_i \prec_\sigma \lambda_j$ implies that $\lambda_i$ is ranked worse than $\lambda_j$ although $\lambda_i$ is relevant while $\lambda_j$ is irrelevant ($y_i > y_j$). The last equation can equivalently be rewritten~\cite{nguyen2021multilabel} as follows
\begin{equation}
\ell_R(\sigma, \vect{y}) =
	\sum_{1\leq i < j \leq m}
		\mathbbm{1}_{(y_{\sigma(i)} = 0 
					\wedge y_{\sigma(j)} = 1)}.
\end{equation}

In the case of precise probabilities, it is also useful to recall that the optimal prediction for the ranking loss~\cite{dembczynski2012label} is the one $\hat{\sigma}$ sorting the labels $\lambda_i$, $i \in [m] :=\{1, \ldots, m\}$, in decreasing order of the probabilities $P_\newinstance(Y_{\hat{\sigma}(i)}=1)$, that is 
\begin{equation}\label{eq:optimal_rank}
	P_\newinstance(Y_{\hat{\sigma}(1)}=1)\geq 
	P_\newinstance(Y_{\hat{\sigma}(2)}=1)\geq 
	\dots\geq 
	P_\newinstance(Y_{\hat{\sigma}(m)}=1).
\end{equation}

When considering a set $\credal$ of distributions, one is immediately tempted to adopt the partial order obtained of $\hat{\sigma}^*_{\ell_R,\credal}$ such that 
\begin{equation}\label{eq:approxHamIP}
	\lambda_i \succ_{\hat{\sigma}^*_{\ell_R,\credal}} \lambda_j
	\iff \underline{P}_\newinstance(Y_i=1) > \overline{P}_\newinstance(Y_j=1).
\end{equation}
It has however been proven that $\hat{\sigma}^*_{\ell_R,\credal}$ is in general an outer-approximation of $\hat{\mathbb{R}}^{M}_{\ell_R,\credal}$, thus only providing a quick heuristic to get an approximate answer~\cite{destercke2015multilabel}.

In the next sections, we study the problem of providing exact skeptic inferences, first for any possible probability set $\mathcal{P}$ and then for the specific case where $\mathcal{P}$ is built from marginal models on each label, the latter corresponding to binary relevance approaches.

\subsection{General case}
In this section, we demonstrate that for the ranking loss, we can use inference procedures that are much more efficient than an exhaustive, naive enumeration.

Let us first simplify the expression of the expected value.
\begin{lemma}\label{prop:expcond}
    In the case of the ranking loss and given $\sigma_1,\sigma_2\in\outrank$, we have
    \begin{align}\label{eq:expcond}
	&\mathbb{E}\left[ \ell_R(\sigma_2, \cdot) - 
		\ell_R(\sigma_1, \cdot) | X=\insta \right] =
	\sum_{1\leq i<j\leq m}
	P_{\newinstance}(Y_{\sigma_2(i)}=0,Y_{\sigma_2(j)}=1)-  
	P_{\newinstance}(Y_{\sigma_1(i)}=0,Y_{\sigma_1(j)}=1) 
	\end{align} 
\end{lemma}

The next proposition shows that this expression can be leveraged to perform the maximality check of Equation \eqref{eq:compaIP} on a limited number of pairwise label preferences.

\begin{proposition}\label{prop:reduction}
	For a given set $\setindices_{\sigma_1\neq\sigma_2}$ of indices defined as 
	$$\setindices_{\sigma_1\neq\sigma_2}=\left\{(i,j) \middle| \lambda_i \succ_{\sigma_1} \lambda_j \text{~and~} \lambda_j \succ_{\sigma_2} \lambda_i \right\},$$


we can rewrite the \hyperref[eq:compaIP]{maximality criterion} as follows
\begin{align}\label{eq:maxireduction}
	\sigma_1 \sqsupset_{\lossrank, \credal} \sigma_2\iff\inf_{P_\insta\in\credal} 
		\hspace{-8mm}
		\sum_{\substack{\mbox{}\vspace{1mm}\\\hspace{6mm}
			(i,j)\in\setindices_{\sigma_1\neq\sigma_2}}} 
		\hspace{-8mm}
		P_{\newinstance}(Y_{i}=1) -  
		P_{\newinstance}(Y_{j}=1)>0
\end{align}
\end{proposition}
Proposition \ref{prop:reduction} amounts to saying that it is only necessary to verify all different disagreements between two ranking solutions (e.g., $\lambda_i\succ_{\sigma_1}\lambda_j$ versus $\lambda_j\succ_{\sigma_2}\lambda_i$). Let us illustrate this matter in the next example.

\begin{example}
Let us take the rankings $\sigma_1$ and $\sigma_2$ already defined in Example~\ref{exa:setdisagreements} and remind the set of pairwise indices of those disagreements between $\sigma_1$ and $\sigma_2$:
\begin{align*}
	\setindices_{\sigma_1\neq\sigma_2}=\{(1,2), (3, 2), (4, 2)\}.
\end{align*}
Given a credal set $\credal$ and the ranking loss $\lossrank$, we can apply the maximality criterion such that $\sigma_1$ is preferred to $\sigma_2$, that is,
\begin{align}\label{eq:exrankings}
	\lambda_1\succ_{\sigma_1}
	\lambda_3\succ_{\sigma_1}
	\lambda_4\succ_{\sigma_1}
	\lambda_2
	{~~~\sqsupset_{\ell_R,\credal}~~~}
	\lambda_2\succ_{\sigma_2} 
	\lambda_1\succ_{\sigma_2} 
	\lambda_3\succ_{\sigma_2}  
	\lambda_4,
\end{align}
if and only if it verifies Equation~\eqref{eq:maxireduction}.

In what follows, we show how it is possible to build a finite set of different rankings $\sigma_*$ that verify the set $\setindices_{\sigma_1\neq\sigma_2}$. To do that, we start by building the set of pairwise label preferences based on the set $\setindices_{\sigma_1\neq\sigma_2}$
\begin{align*}
\mathcal{N}_{\sigma_*} = \{ \lambda_1 \succ_{\sigma_*} \lambda_2,~\lambda_3 \succ_{\sigma_*} \lambda_2,~\lambda_4 \succ_{\sigma_*} \lambda_2 \},
\end{align*}
and, of course, we ignore the preference $\lambda_3\succ\lambda_4$ since it does exist in both rankings $\succ_{\sigma_1}$ and $\succ_{\sigma_2}$, besides the order of its labels  (i.e. $\lambda_3\succ\lambda_4$ or $\lambda_4\succ\lambda_3)$ is indifferent as long as they belong to both rankings. Note that $\sigma_1$ is a super-set of $\sigma_*$.

Using the set $\mathcal{N}_{\sigma_*}$, we can easily build a set of ranking (or a strict total order) on the basis of $\mathcal{N}_{\sigma_*}\cup \{\lambda_3\succ_{\sigma_*}\lambda_4\}$ or  $\mathcal{N}_{\sigma_*}\cup \{\lambda_4\succ_{\sigma_*}\lambda_3\}$ which can verify the maximality criterion of Equation~\eqref{eq:exrankings}:
\begin{align*}
	\lambda_1\succ_{\sigma_3}\lambda_4\succ_{\sigma_3}
	\lambda_3\succ_{\sigma_3}\lambda_2
	&{~~\sqsupset_{\ell_R,\credal}~~}
	\lambda_2\succ_{\sigma_4}\lambda_1\succ_{\sigma_4} 
	\lambda_4\succ_{\sigma_4}\lambda_3,\\
	\lambda_3\succ_{\sigma_5}\lambda_1\succ_{\sigma_5}
	\lambda_4\succ_{\sigma_5}\lambda_2
	&{~~\sqsupset_{\ell_R,\credal}~~}
	\lambda_2\succ_{\sigma_6}\lambda_3\succ_{\sigma_6} 
	\lambda_1\succ_{\sigma_6}\lambda_4,\\
	\lambda_4\succ_{\sigma_7}\lambda_1\succ_{\sigma_7}
	\lambda_3\succ_{\sigma_7}\lambda_2
	&{~~\sqsupset_{\ell_R,\credal}~~}
	\lambda_2\succ_{\sigma_8}\lambda_4\succ_{\sigma_8} 
	\lambda_1\succ_{\sigma_8}\lambda_3,\\
	\lambda_3\succ_{\sigma_9}\lambda_4\succ_{\sigma_9}
	\lambda_1\succ_{\sigma_9}\lambda_2
	&{~~\sqsupset_{\ell_R,\credal}~~}
   \lambda_2\succ_{\sigma_{10}}\lambda_3\succ_{\sigma_{10}} 
	\lambda_4\succ_{\sigma_{10}}\lambda_1,\\
   \lambda_4\succ_{\sigma_{11}}\lambda_3\succ_{\sigma_{11}}
	\lambda_1\succ_{\sigma_{11}}\lambda_2
	&{~~\sqsupset_{\ell_R,\credal}~~}
   \lambda_2\succ_{\sigma_{12}}\lambda_4\succ_{\sigma_{12}} 
	\lambda_3\succ_{\sigma_{12}}\lambda_1.
\end{align*}
Finally, we can also note that using the set $\setindices_{\sigma_1\neq\sigma_2}$, a single checking of Equation~\eqref{eq:compaIP} is required instead of six (i.e. $\sigma_1\sqsupset_{\ell_R,\credal}\sigma_2, \sigma_3\sqsupset_{\ell_R,\credal}\sigma_4,\dots,\sigma_{11}\sqsupset_{\ell_R,\credal}\sigma_{12}$), besides if Equation~\eqref{eq:exrankings} verifies the maximality criterion, so $\sigma_*\in\hat{\mathbb{R}}^{M}_{\ell_R,\credal}$ is a dominant solution.
\end{example}

At present, it is evident that we just need to check all the different combinations of pairwise label preferences built from the set of labels $\outspace$. To do this, we propose the procedure \emph{``checkDisagreement''} of Algorithm~\ref{alg:ranking}
which create all different combinations\footnote{Note that the procedure ``checkDisagreement'' does not allow to build an combination of opposite preferences, for instance: $\setindices_* = \{(1,2), (2,1) \}$, since with such a set, it is impossible to create a (strictly total order) ranking solution $\sigma_*$} in a recursive way by using the set 
\begin{align}
	\mathcal{W}_{\outspace} = \left\{ \lambda_i \succ_{\sigma_*} \lambda_j
		 \middle| \lambda_i, \lambda_j \in \outspace, i\neq j \right\}, 
\end{align}
which contains all the different pairwise label preferences that can be created from the set $\outspace$, and besides $|\mathcal{W}_{\outspace}| = m(m-1)$. Then, we can verify if each combination verifies the maximality criterion by applying Equation~\eqref{eq:maxireduction}.
\newcommand{\pref}[2]{\lambda_{#1} \succ_{\sigma_*} \lambda_{#2}}
\begin{example}
	Given the set of labels $\mathcal{K}=\{\lambda_1, \lambda_2, \lambda_3, \lambda_4\}$, we can create the set of pairwise label preference 
	\begin{align}
	\mathcal{W}_{\outspace} = 
	 \begin{Bmatrix}
	  \cdot &\pref{1}{2} & \pref{1}{3} & \pref{1}{4}\\
	  \pref{2}{1} & \cdot & \pref{2}{3} & \pref{2}{4} \\
	  \pref{3}{1} & \pref{3}{2} & \cdot & \pref{3}{4} \\
	  \pref{4}{1} & \pref{4}{2} & \pref{4}{3} & \cdot 
	 \end{Bmatrix},
	\end{align}
	and we can also build a combination of preferences from $\mathcal{W}_{\outspace}$, so that such a set can be used as a set of disagreement indices to check the maximality criterion of Equation~\eqref{eq:maxireduction}:
	$$\mathcal{N}_{\sigma_*}=\left\{ \pref{2}{1},~\pref{3}{2},~\pref{4}{3} \right\} \iff \setindices_{*}=\{(2,1), (3, 2), (4, 3)\}.$$
\end{example}

\begin{figure}[!ht]
    \centering
	\begin{tikzpicture}[scale=0.75]
	\begin{axis}[
		xlabel=$m$,ylabel=\# of Evaluation needed in $\log_{10}(\cdot)$,
		                xmin=2, xmax=11, samples=17, xtick={2,3,4,...,11},  
		                legend style={at={(0.5,-0.15)},
      anchor=north, legend columns=-1},]
	\addplot[mark=*,black, domain=0:16] {ln(factorial(x)^2)/ln(10)};
	\addplot[mark=square*,red, domain=0:16] {ln(factorial(x)^1.8)/ln(10)};
	\addplot[mark=triangle*, blue] coordinates{(2, 0.301030) (3, 1.255273) (4, 2.365488) (5, 3.655138) (6, 5.086452) (7, 6.626479) (8, 8.253262) (9, 9.951963)(10, 11.71203) (11, 13.52564)};
	\legend{Naive, Outer-complexity ,Algorithm~\ref{alg:ranking}};
	\end{axis}%
	\end{tikzpicture}%
    \caption{Comparison of Algorithm~\ref{alg:ranking} with naive enumeration.}
    \label{fig:compaalgonaive}
\end{figure}
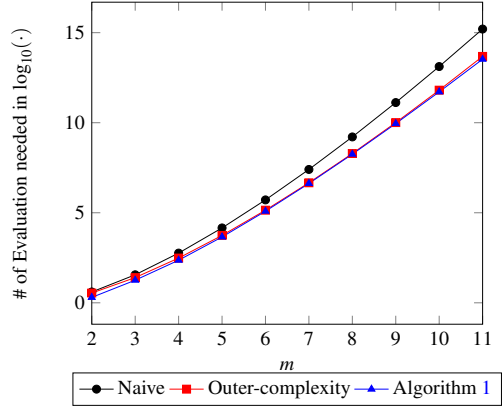

Proposition~\ref{prop:reduction} and Algorithm~\ref{alg:ranking} therefore allow us to find $\hat{\mathbb{R}}^{M}_{\ell_R,\credal}$. The following conjecture provides at first glance of the time complexity of Algorithm~\ref{alg:ranking}, and in Figure~\ref{fig:compaalgonaive}, we plots three different curves as a function of the number $m$ of labels: (1) the naive version performing all verifications, (2) outer-approximation proposed by Conjecture~\ref{conj:complexity}, and (3) the real number of verifications obtained with Algorithm~\ref{alg:ranking}.
\begin{conjecture}\label{conj:complexity}
Algorithm~\ref{alg:ranking} has to perform less than $m!^{1.8}$ computations, and its outer-complexity is in $\mathcal{O}(m!^{1.8})$
\end{conjecture}

\newcommand{\myalgorithm}{%
\begin{algorithm}[ht]
  \SetKw{KwBy}{by}
  \DontPrintSemicolon
  \caption{Maximal solutions under Ranking loss and a general set	distributions}\label{alg:ranking}
  \LinesNumbered
  \SetKwFunction{main}{main}
  \SetKwFunction{maximality}{maximality}
  \SetKwFunction{addpref}{addPreference}
  \SetKwFunction{genSetsRanks}{checkDisagreement}
  \KwData{$\mathscr{P}$ (convex set of distributions)}
  \KwResult{$\hat{\mathbb{R}}^{M}_{\ell_R,\credal}$ (set of undominated  solutions)}
  \SetKwProg{myproc}{Procedure}{}{end}
  \myproc{\maximality{$\setindices_{\sigma_1\neq\sigma_2}$}}{
  	\lIf{${\small 
  		\inf\limits_{P_\insta\in\credal} 
		\hspace{-6mm}
		\sum\limits_{\substack{\mbox{}\vspace{1mm}\\\hspace{6mm}
			(i,j)\in\setindices_{\sigma_1\neq\sigma_2}}}
		\hspace{-6mm}
		P_{\newinstance}(Y_{i}=1)-
		P_{\newinstance}(Y_{j}=1)>0}$}
	{$\hat{\mathbb{R}}^{M}_{\ell_R,\credal} = 
    \hat{\mathbb{R}}^{M}_{\ell_R,\credal} \cup \sigma_*$}
  }{}
  \myproc{\genSetsRanks{l, $\mathcal{W}$, $\setindices$}}{
  	\uIf{$l == 1$}{
  		\lFor{$(\lambda_i \succ \lambda_j) \in \mathcal{W}$}{
    		\maximality{$\setindices\cup(\lambda_i \succ \lambda_j)$}
    	}
  	}
  	\Else{
    	\For{$(\lambda_i \succ \lambda_j) \in \mathcal{W}$}{
    		$\mathcal{W}_1 = \mathcal{W} \backslash 
    			\{(\lambda_i \succ \lambda_j) \cup 
    			(\lambda_j \succ \lambda_i)\}$\;
    		\BlankLine\vspace{-1mm}
    		$\setindices_1 = \setindices 
    			\cup (\lambda_i \succ \lambda_j)$\;
    		\BlankLine\vspace{-1mm}
    		\maximality{$\setindices_1$}\;
    		\BlankLine\vspace{-1mm}
    		\genSetsRanks{$l-1$,$\mathcal{W}_1$,$\setindices_1$}\;
    		\BlankLine\vspace{-1mm}
    		$\mathcal{W} = \mathcal{W}\backslash (\lambda_i \succ \lambda_j)$
    	}
	}
  }{}
  
  \SetKwProg{myalg}{Algorithm}{}{end}
  \myalg{\main{}}{
   	\genSetsRanks{$m-1$, $\mathcal{W}_{\outspace}$, $\emptyset$}\;
   	\BlankLine\vspace{-2mm}
    \KwRet $\hat{\mathbb{R}}^{M}_{\ell_R,\credal}$
  }
\end{algorithm}}
\myalgorithm

Conjecture~\ref{conj:complexity} tells us that, in the case of ranking loss, finding $\hat{\mathbb{R}}^{M}_{\ell,\credal}$ is much harder than in the case of Hamming loss~\cite[Prop. 3]{yonseb2020hammingmlc} and can be done almost below its outer-complexity $\mathscr{O}(m!^{1.8})$ using Algorithm~\ref{alg:ranking}. Furthermore, compared to the naive procedure, the new algorithm drastically reduces the number of comparisons. For instance; for $m=7$, the naive procedure needs to perform $1'625'702'400$ comparisons, but Algorithm~\ref{alg:ranking} only $179'168'816$, which is approximately $11\%$ (or a ninth part) of the naive version. In addition, the number of comparisons decreases as long as $m$ increases (e.g. for $m=11$, it need only $\sim2\%$ of comparisons of the naive version). On the other hand, it is far from being an optimal procedure, since even a single skeptical inference would need a significant processing time on modern computers.

\subsection{Inference using an imprecise probabilistic tree model}

The hardness of computing Equation~\eqref{eq:maxireduction} may highly depend on the imprecise probabilistic model used. That is why, we choose in this paper to use an  \emph{imprecise probabilistic {\bf tree} model} (IPT model) which has widely been studied in the context of sets of probabilities by \citet{hermans2009imprecise}, and also used to make skeptical inferences in multi-label problems \cite{alarcon2021distributionally,yonseb2020hammingmlc}. 

Computing Equation~\eqref{eq:maxireduction} in an IPT model cannot be done directly, since the IPT model needs iteratively compute lower expectations from the leaf to root of the imprecise tree model (for more details we refer to \cite[\S. 4.1]{hermans2009imprecise, yonseb2020hammingmlc}). Thus, in the next proposition, we propose another way to come down it through the subtraction of two expected weighted partial Hamming losses. We will define the weighted partial Hamming loss between a partial binary vector $\vect{a}_{\mathcal{U}}$, in which the values of this vector are restricted to elements indexed in the set $\mathcal{U}\subseteq\setn{m}$ of indices, and an observation $\vect{y}$ as follows
\begin{equation}\label{eq:partial_ham_loss}
\ell_{H\vect{w}}^*(\vect{a}_\mathcal{U},\bm{y})=\sum_{i \in \mathcal{U}} w_i \indicator{a_i \neq y_i},
\end{equation} 
where $w_i$ is the weight given to the label $y_i, i\in\mathcal{U}$. 

\begin{proposition}\label{prop:ranktohamming}
For a given set $\setindices_{\sigma_1\neq\sigma_2}$ of disagreements obtained from ranking $\sigma_1$ and $\sigma_2$, Equation~\eqref{eq:maxireduction} can be rewritten as follows
\begin{align}\label{eq:ranktohamming}
	\inf_{P_\insta\in\credal}\hspace{-2mm}
	\sum_{\substack{\mbox{}\vspace{1mm}\\
			(i,j)\in\setindices_{\sigma_1\neq\sigma_2}}} 
		\hspace{-4mm}
		P_{\newinstance}(Y_{i}=1) -  
		P_{\newinstance}(Y_{j}=1) = 
	 \inf_{P_\insta\in\credal}
	 \mathbb{E}\left[
		\ell_{H\vect{w}}^*(\vect{0}_{\mathcal{U}^i}, \cdot) -
		\ell_{H\vect{w}}^*(\vect{0}_{\mathcal{U}^j}, \cdot) 
	\right], 
\end{align}
where $\ell_{H\vect{w}}^*(\cdot, \cdot)$ is the weighted partial Hamming loss, $\vect{0}_{\mathcal{U}^*}$ is a partial vector of zero values, and $\mathcal{U}^i$ and $\mathcal{U}^j$ are sets of unique or distinct indices obtained from $(i, j)\in\setindices_{\sigma_1\neq\sigma_2}$ and defined as follows
\begin{align*}
	\mathcal{U}^i &= \{ i ~|~ (i, \cdot) \in \setindices_{\sigma_1\neq\sigma_2} \} \quad\text{such that}\quad \forall (i, \cdot)\in\setindices_{\sigma_1\neq\sigma_2},~~ \text{so}~~ |i \in \mathcal{U}^i| = 1,  \\
	\mathcal{U}^j &= \{ j ~|~ (\cdot, j) \in \setindices_{\sigma_1\neq\sigma_2} \} \quad\text{such that}\quad \forall (\cdot, j)\in\setindices_{\sigma_1\neq\sigma_2},~~ \text{so}~~ |j \in \mathcal{U}^j| = 1.
\end{align*}
Besides, the weighted vector $\vect{w}^{\mathcal{U}^i}=(w_1, \dots, w_m)$ used in $\ell_{H\vect{w}}^*(\mathcal{U}^i, \cdot)$, in which records the number of times each index $i$ is repeated, is defined as follows: 
\begin{align*}
	w_i = \#\left\{(i, \cdot) \in \setindices_{\sigma_1\neq\sigma_2}~\middle|~ i \in \mathcal{U}^i\right\}
\end{align*}
and the weighted vector $\vect{w}^{\mathcal{U}^j}=(w_1, \dots, w_m)$ used in $\ell_{H\vect{w}}^*(\mathcal{U}^j, \cdot)$, similarly
\begin{align*}
	w_j =\#\left\{(\cdot, j) \in \setindices_{\sigma_1\neq\sigma_2}~\middle|~ j \in \mathcal{U}^j\right\}
\end{align*}
\end{proposition}

Let us illustrate Proposition~\ref{prop:ranktohamming} by making use of the imprecise probabilistic tree models in the next example.
\begin{example}\label{ex:ranking}
	Consider the imprecise probabilistic tree developed in Figure~\ref{fig:imprecisetree} defined over $\mathcal{Y}= \{0, 1\}^3$ describing an imprecise joint distribution $\hat{\credal}$ over three labels (i.e. the set of labels $\outspace = \{\lambda_1, \lambda_2, \lambda_3\}$). Here, we  will calculate the $\hat{\mathbb{R}}^{M}_{\ell_R,\hat{\credal}}$ by using Algorithm~\ref{alg:ranking} and Proposition~\ref{prop:ranktohamming}.
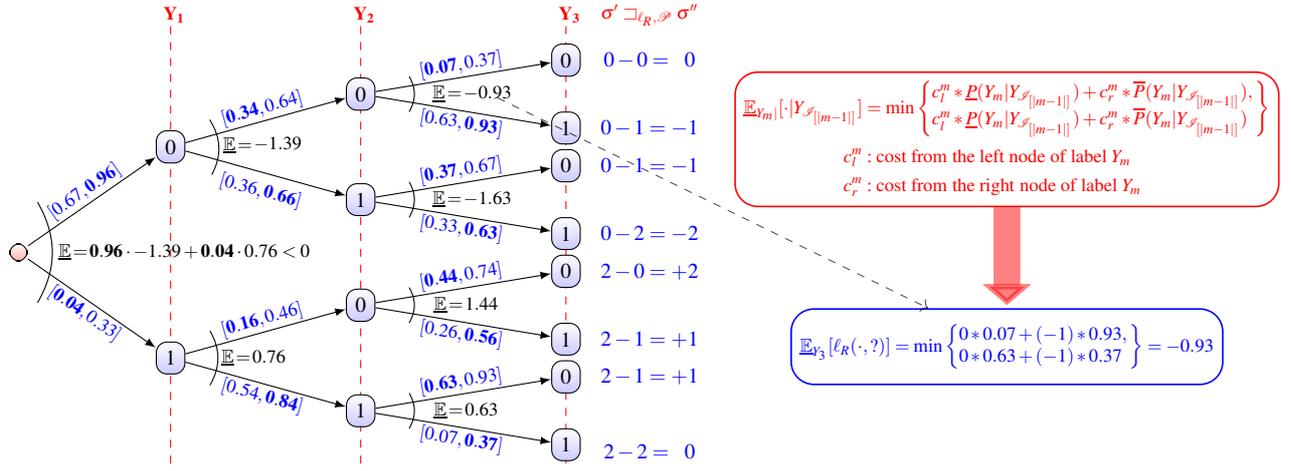
\begin{figure}[H]
	\centering
	\hspace{-6mm}\input{eg_improb_loss_tree.tex}
	\caption{Imprecise probabilistic trees}
	\label{fig:imprecisetree}
\end{figure}

Before applying Algorithm~\ref{alg:ranking}, we first build the set $\mathscr{W}_\mathscr{K}$ of pairwise preferences over $\outspace$ as follows:
\begin{equation}
\mathscr{W}_\mathscr{K}=\begin{Bmatrix}
	\cdot &\lambda_1\succ\lambda_2, & \lambda_1\succ\lambda_3, \\
	\lambda_2\succ\lambda_1 & \cdot & \lambda_2\succ\lambda_3, \\
	\lambda_3\succ\lambda_1, & \lambda_3\succ\lambda_2 & \cdot
\end{Bmatrix},
\end{equation}
and then, by using $\mathscr{W}_\mathscr{K}$, we can apply Algorithm~\ref{alg:ranking} which builds recursively all the sets of disagreements (or sets of pairwise indices $\setindices_{\sigma'\neq\sigma''}$ of all the different couple of rankings on which have oposite label preferences, see Example~\ref{exa:setdisagreements}) by calling the procedure ``checkDisagreement''. It can easily verify that it is possible to build $18$ sets of disagreement indices using the set $\mathscr{W}_\mathscr{K}$ and the procedure ``checkDisagreement'', as follows:
\begin{align}
	\begin{Bmatrix}
		\begin{matrix}
			&\{(1, 2)\}, \\
			&\{(1, 2), (1, 3)\}, \\
			&\{(1, 2), (2, 3)\},\\
			&\{(1, 2),(3, 1)\},\\
			&\{(1, 2), (3, 2)\},\\
		\end{matrix} &
		\begin{matrix} 
			&\{(1, 3)\},\\ 
			&\{(1, 3), (2, 1)\}, \\
			&\{(1, 3), (2, 3)\}, \\
		    &\{(1, 3), (3, 2)\},\\	
		\end{matrix} &
		\begin{matrix}
			&\{(2, 1)\}, \\
			&\{(2, 1), (2, 3)\}, \\
			&\{(2, 1), (3, 1)\}, \\
			&\{(2, 1), (3, 2)\},\\			
		\end{matrix} &
		\begin{matrix}
			&\{(2, 3)\},\\
			&\{(2, 3), (3, 1)\},\\
			&\{(3, 1)\},\\
			&\{(3, 1), (3, 2)\},\\
			&\{(3, 2)\}			
		\end{matrix}
	\end{Bmatrix}
\end{align}

While each set of disagreement indices is created recursively, Algorithm~\ref{alg:ranking} verifies if each one of them is a solution of $\hat{\mathbb{R}}^{M}_{\ell_R,\hat{\credal}}$. For instance, in Figure~\ref{fig:imprecisetree}, we verify if the set of disagreement indices $\setindices_{\sigma'\neq\sigma''}=\{(1, 2), (1, 3)\}$, which produces the following rankings 
\begin{align}\label{eq:exranksols}
	&\setindices_{\sigma'\neq\sigma''} \iff 
	\begin{aligned}
		\sigma'\qquad&\sqsupset_{\lossrank, \hat{\credal}}
		\qquad\sigma''\\
		\lambda_1\succ{\color{blue}\lambda_2\succ\lambda_3}
			&\sqsupset_{\lossrank, \hat{\credal}}
		{\color{blue}\lambda_2\succ\lambda_3}\succ\lambda_1\\
		\lambda_1\succ{\color{blue}\lambda_3\succ\lambda_2}
			&\sqsupset_{\lossrank, \hat{\credal}}
		{\color{blue}\lambda_3\succ\lambda_2}\succ\lambda_1,
	\end{aligned}
\end{align}
satisfy the maximality criterion of Equation~\eqref{eq:maxireduction}. To do this, we calculate weighted partial Hamming losses of each set of unique indices, i.e. $\mathcal{U}^i=\{1\}$ and $\mathcal{U}^j=\{2, 3\}$ (with the weighted vectors $\vect{w}^{\mathcal{U}^i}=(1, 0, 0)$ and $\vect{w}^{\mathcal{U}^j}=(0, 1, 1)$), as follows 
\begin{center}
\begin{tabular}{c|cc}
\hline $\vect{y}$ & $\ell_{Hw}^*(\vect{0}_{\mathcal{U}^i}, \cdot)$  & 
	$\ell_{Hw}^*(\vect{0}_{\mathcal{U}^j}, \cdot)$ \\\hline
$(0,0,0)$ & $0$ & $0$ \\
$(0,0,1)$ & $0$ & $1$ \\
$(0,1,0)$ & $0$ & $1$ \\
$(0,1,1)$ & $0$ & $2$ \\
$(0,0,0)$ & $2$ & $0$ \\
$(0,0,1)$ & $2$ & $1$ \\
$(0,1,0)$ & $2$ & $1$ \\
$(0,1,1)$ & $2$ & $2$ \\
\hline
\end{tabular}
\end{center}	
where $\cdot$ should be replaced by an observed output $\vect{y}$. Then, we recursively compute the infimum expectation of Equation~\eqref{eq:ranktohamming} using the law of iterated lower expectation described by \citet{hermans2009imprecise} and \citet[\S 4.1]{yonseb2020hammingmlc}. Finally, the ranking solutions of Equation~\eqref{eq:exranksols} built from the set of indices $\setindices_{\sigma'\neq\sigma''}$ does not belong to $\hat{\mathbb{R}}^{M}_{\ell_R,\hat{\credal}}$, as the infimum expectation is negative, i.e. $\lexpe\!=\!\mathbf{0.96} \cdot -1.39 + \mathbf{0.04} \cdot 0.76 < 0$ 
\end{example}

Note that, in Example~\ref{ex:ranking}, it does not need to verify all different rankings of Equation~\eqref{eq:exranksols}, but just one of them, or more specifically, the set of disagreement indices $\setindices_{\sigma'\neq\sigma''}$. Thus, it reduces as much as possible the number of solutions which should verify the maximality criterion.

So far we prove that under the maximality principle, it is not enough to consider marginal probabilities in order to get set-valued optimal predictions. Yet, in what follows, we show that, on special credal sets, knowledge of the marginal probability is enough to determinate the optimal set $\hat{\mathbb{R}}^M_{\ell,\credal}$. 

\newcommand{\setindicesind}{\mathscr{J}}
\subsection{The case of label independence}
In this section, we assume (conditional) independence of label probabilities in the sense that \citep{dembczynski2012label}
\begin{align}\label{eq:independence}
P_{\newinstance}(\vect{y}) =\prod_{i=1}^m P_{\newinstance}(Y_{i}=y_i)
\, , \forall \vect{y} \in \mathcal{Y}\, .
\end{align}

It is important to keep in mind that learning the optimal prediction in a MLC problem, even in the precise probability setting, can be harder under the general assumption of label dependence. On the other hand, in the case of the label independence assumption \eqref{eq:independence} becomes treatable, even with non-decomposable losses including the F-measure, the Jaccard measure and the subset $0/1$ loss. In the related work of MLC with partial abstention, \citet{nguyen2021multilabel} also reduced the investigation on the label independence assumption when learning the optimal partial prediction of the Rank loss and F-measure.    

In the following, we show that even under the label independence assumption the problem of learning the optimal predictions of the E-admissibility and Maximality principle can be done efficiently, but of course it is not obvious. 

Beside of the label independence assumption of Equation \eqref{eq:independence}, we assume the credal (interval) marginals are made available by some imprecise multi-label classifier, i.e., $\forall i \in [m]$, we have
\begin{align}\label{eq:credal_marginals}
    P_{\newinstance}(Y_{i}=1) \in \credal_i := \left[ \underline{P}_{\newinstance}(Y_{i}=1)\, , \overline{P}_{\newinstance}(Y_{i}=1)\right]  \, .
\end{align}
The credal marginals of Equation \eqref{eq:credal_marginals} and the label independence assumption of Equation \eqref{eq:independence} allow us to build the following credal set of the joint probability distribution
\begin{align}\label{eq:credal_joint_independence}
  \credalind := \left\{P_{\newinstance} \, \middle| \,  P_{\newinstance}(Y_{i}=1) \in \credal_i, P_{\newinstance}(\vect{y}) =\prod_{i=1}^m P_{\newinstance}(Y_{i}=y_i)\right\} \, ,  
\end{align}
because, for any $\left(P_{\newinstance}(Y_{1}=1), \ldots , P_{\newinstance}(Y_{m}=1) \right) \in \credal_1 \times \ldots \times \credal_m$, we can show that 
\begin{align*}
     \sum_{\vect{y} \in \mathcal{Y}}P_{\newinstance}(\vect{y}) =\sum_{\vect{y} \in \mathcal{Y}} \prod_{i=1}^m P_{\newinstance}(Y_{i}=y_i) =   1 \, .
\end{align*}
\begin{lemma}\label{lem:pairwise_difference}
	 Let $\sigma^{-1}(i)$ be the position of label $i$-th in the rank $\sigma$. Given the set of indices $\setindicesind^{\sigma_2\neq\sigma_1}$ obtained from two ranking solutions $\sigma_1$ and $\sigma_2$, and defined as follows   
	 $$\setindicesind^{\sigma_2\neq\sigma_1}=\left\{i \middle| \sigma^{-1}_2(i) \neq \sigma^{-1}_1(i) \right\},$$ 
	then under the label independence assumption of Equation \eqref{eq:independence} and the imprecise joint probability distribution of Equation \eqref{eq:credal_joint_independence}, we have that 
\begin{align}\label{eq:pairwise_difference}
\sigma_1 \sqsupset_{\lossrank, \credalind} \sigma_2 \iff
&\inf_{P_{\newinstance} \in\credalind}\sum_{
		i\in\setindicesind_{\sigma_1\neq\sigma_2}} 
	P_{\newinstance}(Y_{i}=1) 
	(\sigma^{-1}_2(i) - \sigma^{-1}_1(i)) >0 \nonumber\\
\iff 
&\sum_{
		i\in\setindicesind_{\sigma_1\neq\sigma_2}} 
	\inf_{P_{\newinstance}(Y_{i}=1) \in\credal_i}P_{\newinstance}(Y_{i}=1) 
	(\sigma^{-1}_2(i) - \sigma^{-1}_1(i)) >0\nonumber \\
\iff	 
&\sum_{
		i\in\setindicesind_{\sigma_1\neq\sigma_2}} 
	P_{\newinstance}^*(Y_{i}=1) 
	(\sigma^{-1}_2(i) - \sigma^{-1}_1(i)) >0	\, ,
	\end{align}
where
\begin{align}
 	P_{\newinstance}^*(Y_{i}=1) 
 	=
 	\begin{cases}
 	 	\underline{P}_{\newinstance}(Y_{i}=1) \text{ if }  \sigma^{-1}_2(i) > \sigma^{-1}_1(i) \,,\\
 	 	\overline{P}_{\newinstance}(Y_{i}=1) \text{ otherwise. } \\
 	\end{cases}
\end{align}
\end{lemma}

Lemma \ref{lem:pairwise_difference} amounts to saying that it is only necessary to have knowledge about the ranks of labels $Y_i$ into the two ranking $\sigma_*$ to compute the correct bound probability $P_{\newinstance}^*(Y_{i}=1)$. 

In what follows, Proposition~\ref{prop:imprecise_ranks} shows that the possible ranks which can be assigned to each label belong to an interval. Moreover, the intervals can be determined efficiently using the upper and lower marginal probabilities.
\begin{proposition}\label{prop:imprecise_ranks}
Assuming that, for any query instance $\newinstance$, the joint probability $P$ over $\mathcal{Y}$ and its imprecise extension $\credalind$ are defined as Equation \eqref{eq:independence} and \eqref{eq:credal_joint_independence}, respectively. Let $\zeta$ be a $m \times m$ matrix defined as 
\begin{align}\label{eq:zeta_matrix}
\zeta_{i,j} = 
\begin{cases}
1 \text{ if } \underline{P}_{\newinstance}(y_i=1)  > \overline{P}_{\newinstance}(y_j=1) \, , i \neq j   \\
1 \text{ if } i = j \, ,   \\
0 \text{ otherwise. }
\end{cases}
\end{align}
Each label $\lambda_i \in \mathcal{Y}$ can be associated to an imprecise rank $[\underline{\sigma}_i, \overline{\sigma}_i]$ with 
\begin{align}
\overline{\sigma}_i & =  m+1 - \sum_{j=1}^m\zeta_{i,j} \, ,\label{eq:upper_rank} \\ \underline{\sigma}_i  & =  \sum_{j=1}^m\zeta_{j,i} \, ,\label{eq:lower_rank}
\end{align}
where $\underline{\sigma}_i$ and $\overline{\sigma}_i$ are respectively the smallest and largest rank that can be given to $\lambda_i$ by any $\hat{\sigma}^{P}_{\ell_R} \in \hat{\mathbb{R}}^E_{\ell_R,\credalind}$.
\end{proposition}

The following example illustrates how to find the imprecise ranks given imprecise marginal probabilities.
\begin{example}\label{exa:example_imprecise_rank}
Let us consider an example where $\mathcal{Y} = \{\lambda_1, \ldots, \lambda_5\}$ and the imprecise marginal probabilities are given in Table \ref{tab:imprecise_probabilites}. 

\begin{table}[H]
\begin{center}
	\caption{Imprecise marginal probabilities}
	\begin{tabular}{c|ccccc}
	&$\lambda_1$ & $\lambda_2$ & $\lambda_3$ & $\lambda_4$ & $\lambda_5$ \\
	\hline
    $\underline{P}_{\newinstance}$ & 0.3 & 0.1  & 0.15 & 0.4 & 0.8 \\
    $\overline{P}_{\newinstance}$  & 0.5 & 0.2  & 0.45 & 0.65 & 0.9
	\end{tabular}
	\label{tab:imprecise_probabilites}
\end{center}
\end{table} 

The corresponding $\zeta$ matrix given in Table~\ref{tab:zeta_matrix} is determined using Equation \eqref{eq:zeta_matrix}.
\begin{table}[H]
\begin{center}

	\caption{The corresponding $\zeta$ matrix}
	\begin{tabular}{c|ccccc|c}
		 & $\lambda_1$ & $\lambda_2$ & $\lambda_3$ & $\lambda_4$ & $\lambda_5$ &$\sum_{j}\zeta_{i,j}$\\
		\hline
	 $\lambda_1$ &1 &1 &0 &0 &0 &2 \\
	 $\lambda_2$ &0 &1 &0 &0 &0 &1\\
	 $\lambda_3$ &0 &0 &1 &0 &0 &1\\
	 $\lambda_4$ &0 &1 &0 &1 &0 &2\\
	 $\lambda_5$ &1 &1 &1 &1 &1 &5\\
	 \hline 
	 $\sum_{j}\zeta_{j,i}$  &2 &4 &2 &2 &1 
	\end{tabular}
	\label{tab:zeta_matrix}
\end{center}
\end{table} 

By applying Equation \eqref{eq:lower_rank} and \eqref{eq:upper_rank}, we can easily compute the imprecise ranks of the training instance $\newinstance$.
\begin{table}[H]
\begin{center}
	\caption{The corresponding imprecise ranks}
	\begin{tabular}{c|ccccc}
	& $\lambda_1$ & $\lambda_2$ & $\lambda_3$ & $\lambda_4$ & $\lambda_5$\\
    \hline
    $\underline{\sigma}$ & 2 & 4 & 2 & 2 & 1 \\
    $\overline{\sigma}$ & 4 & 5 & 5 & 4 & 1 
	\end{tabular}
	\label{tab:imprecise_ranks}
\end{center}
\end{table} 
\end{example}

The following proposition ensures that the optimal predictions of the E-admissibility criterion is identical to the set possible rankings induced from the imprecise ranks of Proposition~\ref{prop:imprecise_ranks}.

\begin{proposition}\label{prop:equality_relation_IE}
Given the imprecise ranks defined in Equation \eqref{eq:upper_rank} and \eqref{eq:lower_rank}, a ranking $\sigma$ is said to be a linear extension of the partial order
\begin{align}
    \sigma^{-1}(i) \in [\underline{\sigma}_i, \overline{\sigma}_i]\, , \forall i \in [m] \, \quad\text{and}\quad 
    \sigma^{-1}(i) \neq \sigma^{-1}(j) \, , \forall i \neq j \, .\label{eq:conditions}
\end{align}
We denote by  
\begin{align}\label{eq:possible_ranking}
 \hat{\mathbb{R}}^{LE}_{\ell,\credalind} = \left\{\sigma \in \outrank ~\middle|~ \sigma ~  \text{s.t. Equation \eqref{eq:conditions}}\right\}.
\end{align}
the set of of linear extensions of the partial order.
Assume the joint probability $P$ over $\mathcal{Y}$ and its imprecise extension $\credal$ are defined as Equation \eqref{eq:independence} and \eqref{eq:credal_joint_independence}, respectively, we have the following equality: 
\begin{align}\label{eq:equality_relation}
   \hat{\mathbb{R}}^{LE}_{\ell,\credalind} =  \hat{\mathbb{R}}^E_{\ell,\credalind} \, .
\end{align}
\end{proposition}

Owing to Proposition \ref{prop:equality_relation_IE}, the optimal solution under E-admissibility can be formulated as a constraint satisfaction problem (CSP) \citep{gent2008generalised} given the imprecise ranks. 
More precisely, we look for all the possible rankings which can derived from the imprecise ranks defined in Equation \eqref{eq:upper_rank} and \eqref{eq:lower_rank}.
 
To complete our investigation for the case of imprecise binary relevance, we show that the optimal predictions of the E-admissibility and Maximality criteria are identical given Equation \eqref{eq:credal_marginals} and \eqref{eq:credal_joint_independence}.

\begin{proposition}\label{prop:equality_relation}
Assuming that, for any query instance $\newinstance$, the joint probability $P$ over $\mathcal{Y}$ and its imprecise extension $\credal$ are defined as Equation \eqref{eq:credal_marginals} and \eqref{eq:credal_joint_independence}, respectively. Then, we have the following equivalence
\begin{align}\label{eq:equality_relation}
   \hat{\mathbb{R}}^E_{\ell_R,\credal} =  \hat{\mathbb{R}}^M_{\ell_R,\credal} \, .
\end{align}
\end{proposition}

\bibliography{bibliography.bib}
\newpage
\appendix 
\section{Supplemental material}

\begin{proof}[\bf Proof of Lemma~\ref{prop:expcond}]
	To simplify notations and readability of the proof, we denote $\sigma^2:=\pi$ and $\sigma^1:=\sigma$.
	
	Let us first develop $\mathbb{E}\left[ \ell_R(\pi, \cdot) - \ell_R(\sigma, \cdot) \right|X=\insta]$:
\begin{align}
	\sum_{\bm y \in \spacelabel} &
		\left(\rank{\pi} - \rank{\sigma} \right)
		P_\insta(Y=\bm y)\\
	\sum_{y_1\in \{0,1\}}\sum_{y_2\in \{0,1\}}\dots\sum_{y_m\in \{0,1\}}
 		&\left( \rank{\pi} - \rank{\sigma} \right)
 			P_\insta(Y=\bm y)
\end{align}

Developing and arranging the sum of the left term inside the brackets with respect to the ranking $\pi$ (the right term can also be treated in a similar way), we obtain
\begin{align}
	\sum_{y_{\pi(1)}\in \{0,1\}}\sum_{y_{\pi(2)}\in \{0,1\}}\dots\sum_{y_{\pi(m)}\in \{0,1\}}
	\rank{\pi} P_x(Y=\bm y)
\end{align}
For two given successive indices $k, s \in \{1,\ldots,m\}$ such that $k<s$, let us consider the rewriting
\begin{align}\label{eq:ksindices}
		\overbrace{\sum_{y_{\pi(1)}\in \{0,1\}}\sum_{y_{\pi(2)}\in \{0,1\}}\dots\sum_{y_{\pi(m)}\in\{0,1\}}}^{m-2}
 		&\left(\sum_{y_{\pi(k)}\in \{0,1\}}
 			\sum_{y_{\pi(s)}\in \{0,1\}}
 			\rank{\pi} P_x(Y=\bm y) \right)
\end{align}

Developing the sum between brackets, we get 
\newcommand{\sumij}{\sum_{\substack{y_{\pi(k)}\in \{0,1\}\\
	y_{\pi(s)}\in \{0,1\}}}}
\newcommand{\sumi}[1]{\sum_{y_{\pi(#1)}\in \{0,1\}}}
\begin{align}
	&=\sumij \left( \ranki{\pi}{1}{2} + \dots + 
	\ranki{\pi}{k}{s}
	+ \dots + \ranki{\pi}{m-1}{m}\right) P_x(Y=\bm y)\nonumber\\
	&= \ranki{\pi}{1}{2} P_x(Y_{\{-k,-s\}}) + \dots + 
	\sumij \ranki{\pi}{k}{s} P_x(Y_{\{k, s\}})
	+ \dots + \ranki{\pi}{m-1}{m} P_x(Y_{\{-k,-s\}})\nonumber\\
	& = \sumij \ranki{\pi}{k}{s} P_x(Y_{\{k, s\}}) + 
	\sum_{\substack{1\leq i< j \leq m\\ (i\neq k, j\neq s)}} 
	\ranki{\pi}{i}{j} P_x(Y_{\{-k,-s\}}) \nonumber\\
	&\qquad + \sumi{k} \left(\sum_{1<i<k}\ranki{\pi}{i}{k} + 
		 \sum_{\substack{k<j\leq m\\j\neq s}}\ranki{\pi}{k}{j} 
		 \right) P_x(Y_{\{-s\}})\nonumber \\
	&\qquad + \sumi{s} \left( \sum_{\substack{1\leq i<s\\i\neq k}}\ranki{\pi}{i}{s} 
		+ \sum_{s<j\leq m}\ranki{\pi}{s}{j} \right) P_x(Y_{\{-k\}}) \label{eq:reductionrank}
\end{align}
where 
\begin{align}
	P_x(Y_{\{k, s\}}) &:= P_x(Y_{\pi(1)}, \dots, Y_{\pi(k)}=y_{\pi(k)}, \dots, Y_{\pi(s)}=y_{\pi(s)}, \dots, Y_{\pi(m)}), \\
	P_x(Y_{\{-k\}})&:=P_x(Y_{\pi(1)}, \dots, Y_{\pi(k-1)}, Y_{\pi(k+1)}, \dots, Y_{\pi(m)}),\\
	P_x(Y_{\{-k,-s\}})&:=P_x(Y_{\pi(1)}, \dots, Y_{\pi(k-1)}, Y_{\pi(k+1)}, \dots, Y_{\pi(s-1)}, Y_{\pi(s+1)}, \dots, Y_{\pi(m)}).
\end{align}
We put back the $(m-1)$ external sums of Equation~\eqref{eq:ksindices} in the first term of Equation~\eqref{eq:reductionrank}, i.e. 
\begin{align}
	&  \overbrace{\sum_{y_{\pi(1)}\in \{0,1\}}\sum_{y_{\pi(2)}\in \{0,1\}}\dots\sum_{y_{\pi(m)}\in\{0,1\}}}^{m-2} 
		\left[ \sumij \ranki{\pi}{k}{s} P_x(Y_{\{k, s\}})\right] = P_x(Y_{\pi(k)}=0, Y_{\pi(s)}=1)
\end{align}
We can apply the same operation shown above to the rest of the left term of Equation~\eqref{eq:ksindices} recursively, i.e. selecting two other successive indices $k', s'$ such that $k'<s'$, and we obtain 
\begin{align}
	\sum_{\bm y \in \spacelabel} \rank{\pi} P_x(Y=\bm y) = \sum_{1\leq i<j\leq m}  
			P_{\newinstance}(Y_{\pi(i)}=0, Y_{\pi(j)}=1) 
\end{align}
By applying the same logic on the ranking $\sigma$, so we finally obtain
\begin{align}
	\sum_{1\leq i<j\leq m}  
		P_{\newinstance}(Y_{\pi(i)}=0, Y_{\pi(j)}=1) -  
		P_{\newinstance}(Y_{\sigma(i)}=0, Y_{\sigma(j)}=1)	
\end{align}
\end{proof}

\begin{proof}[\bf Proof of Proposition~\ref{prop:reduction}]
Let us define the set of pairwise indices $\setindices = \{(i,j)| 1\leq i<j\leq m \}$, and then, we divide it in two exclusive sets of pairwise indices, such that $\setindices=\setindices_{\sigma_2=\sigma_1}\cup\setindices_{\sigma_2\neq\sigma_1}$, where each one is defined as follows
\begin{align}
	\setindices_{\sigma_1\neq\sigma_2}=\left\{(i,j) \middle| \lambda_i \succ_{\sigma_1} \lambda_j \text{~and~} \lambda_j \succ_{\sigma_2} \lambda_i \right\}, \\
	\setindices_{\sigma_1=\sigma_2}=\left\{(i,j) \middle| \lambda_i \succ_{\sigma_1} \lambda_j \text{~and~} \lambda_i \succ_{\sigma_2} \lambda_j \right\}.
\end{align}
\begin{itemize}
	\item The first set of indices $\setindices_{\sigma_1\neq\sigma_2}$ captures those pairwise label preferences on which the ranking $\sigma_1$ and $\sigma_2$ have some disagreement (see Section~\ref{sec:prel}),	
	\item whereas, the second set of indices $\setindices_{\sigma_1=\sigma_2}$ captures those pairwise label preferences on which the rankings $\sigma_1$ and $\sigma_2$ agree on. 
\end{itemize}

It is easy to see that Equation~\eqref{eq:expcond} is cancelled on the set of indices $\setindices_{\sigma^2=\sigma^1}$, and hence, it can be written as follows
\begin{align}
	\sum_{(i,j) \in \setindices_{\sigma'\neq\sigma''}}  
		P_{\newinstance}(Y_{j}=0, Y_{i}=1) -  
		P_{\newinstance}(Y_{i}=0, Y_{j}=1),
\end{align}
adding the terms $P_{\newinstance}(Y_{i}=1, Y_{j}=1)-P_{\newinstance}(Y_{i}=1, Y_{j}=1)$, we get
\begin{align}
	\sum_{(i,j) \in \setindices_{\sigma'\neq\sigma''}}  
		P_{\newinstance}(Y_{i}=1) -  P_{\newinstance}(Y_{j}=1),
\end{align}
and by applying the infimum operator
\begin{align}
	\sigma_1 \sqsupset \sigma_2 \iff \inf_{P\in\credal} \sum_{(i,j) \in \setindices_{\sigma_1\neq\sigma_2}}  
		P_{\newinstance}(Y_{i}=1) -  P_{\newinstance}(Y_{j}=1).
\end{align}
This completes the proof.
\end{proof}

\begin{proof}[\bf Proof of Proposition~\ref{prop:ranktohamming}]
Given the sum inside of Equation~\eqref{eq:maxireduction}, we can rewrite it as follows:
\begin{align}\label{eq:rewritemaxireduction}
	\sum_{(i,j)\in\setindices_{\sigma_1\neq\sigma_2}}
	P_{\newinstance}(Y_{i}=1) -  P_{\newinstance}(Y_{j}=1) 
	= \sum_{(i,\cdot)\in\setindices_{\sigma_1\neq\sigma_2}} P_{\newinstance}(Y_{i}=1)
	 - \sum_{(\cdot,j)\in\setindices_{\sigma_1\neq\sigma_2}} P_{\newinstance}(Y_{j}=1)
\end{align}
As it is possible that some indices $(i, \cdot)$ or $(\cdot, j)$ may repeat several times, we define two sets of distinct indices:
\begin{align*}
	\mathcal{U}^i &= \{ i ~|~ (i, \cdot) \in \setindices_{\sigma_1\neq\sigma_2} \} \quad\text{such that}\quad \forall (i, \cdot)\in\setindices_{\sigma_1\neq\sigma_2},~~ \text{so}~~ |i \in \mathcal{U}^i| = 1,  \\
	\mathcal{U}^j &= \{ j ~|~ (\cdot, j) \in \setindices_{\sigma_1\neq\sigma_2} \} \quad\text{such that}\quad \forall (\cdot, j)\in\setindices_{\sigma_1\neq\sigma_2},~~ \text{so}~~ |j \in \mathcal{U}^j| = 1.
\end{align*}
Besides, we define the weighted vector $\vect{w}^{\mathcal{U}^i}=(w_1, \dots, w_m)$, in which records the number of times each index $i$ is repeated, as follows:
\begin{align*}
	w_i = \#\left\{(i, \cdot) \in \setindices_{\sigma_1\neq\sigma_2}~\middle|~ i \in \mathcal{U}^i\right\}
\end{align*}
and the weighted vector $\vect{w}^{\mathcal{U}^j}=(w_1, \dots, w_m)$, similarly
\begin{align*}
	w_j =\#\left\{(\cdot, j) \in \setindices_{\sigma_1\neq\sigma_2}~\middle|~ j \in \mathcal{U}^j\right\}
\end{align*}

Therefore, by using the set of distinct indices and the weighted vectors instead in Equation \eqref{eq:rewritemaxireduction}, we can rewrite Equation~\eqref{eq:maxireduction} as the expectation of the subtraction of two weighted partial Hamming losses:
\begin{align*}
	\sum_{i\in\mathcal{U}^i}  w_i^{\mathcal{U}^i}~P_{\newinstance}(Y_i=1) -
	\sum_{j\in\mathcal{U}^j}  w_j^{\mathcal{U}^j}~P_{\newinstance}(Y_j=1)
	&=\sum_{y\in\mathcal{Y}} \left[
		\sum_{i\in\mathcal{U}^i}  
		w_i \indicator{y_i \neq 0} P_{\newinstance}(Y=y) -
		\sum_{j\in\mathcal{U}^j}
		w_j \indicator{y_j \neq 0} P_{\newinstance}(Y=y) \right]\\ 
	&=\sum_{y\in\mathcal{Y}} \left[
		\ell_{H\vect{w}}^*(\vect{0}_{\mathcal{U}^i}, y) 
			P_{\newinstance}(Y=y) 	-
		\ell_{H\vect{w}}^*(\vect{0}_{\mathcal{U}^j}, y) 
			P_{\newinstance}(Y=y) \right] \\ 
	&=\underline{\mathbb{E}}\left[
		\ell_{H\vect{w}}^*(\vect{0}_{\mathcal{U}^i}, \cdot) -
		\ell_{H\vect{w}}^*(\vect{0}_{\mathcal{U}^j}, \cdot) 
	\right]
\end{align*}
where $\ell_{H\vect{w}}^*(\cdot, \cdot)$ is defined as the weighted partial Hamming loss and $\vect{0}_{\mathcal{U}^*}$ is a partial vector of zero values.
\end{proof}


\begin{proof}[\bf Proof of Lemma~\ref{lem:pairwise_difference}]
For a labeling $\bm y$, let $r(\bm y) = \sum_{i=1}^m y_i$
be the number of relevant labels, and $c(\bm y) = r(\bm y)(m-r(\bm y))$ the number of relevant/irrelevant label pairs. It can be shown \cite{nguyen2021multilabel} that $c(\bm y)$) is a constant that does not depend on $\sigma$ and 
\begin{align*}
  \mathbb{E}\left[ \ell_R(\sigma, \cdot) \middle| X=\newinstance \right] = \frac{1}{2}\mathbb{E}\left[ \ell_R(\sigma, \cdot) - \ell_R(\overline{\sigma}, \cdot)  \middle| X=\newinstance \right] + \frac{1}{2}\mathbb{E}\left[c(\cdot)  \middle| X=\newinstance \right]   
\end{align*}
with $\overline{\sigma}$ is the reversal of $\sigma$. Furthermore, we have the following relations:
\begin{align*}
    \mathbb{E}\left[ \ell_R(\sigma, \cdot) - \ell_R(\overline{\sigma}, \cdot)  \middle| X=\newinstance \right] &= \sum_{i=1}^m (2i-(m+1))P_{\newinstance}(Y_{\sigma(i)}=1) = \sum_{i=1}^m (2\sigma^{-1}(i)-(m+1))P_{\newinstance}(Y_{i}=1) \, , \\
 \mathbb{E}\left[c(\cdot)  \middle| X=\newinstance \right]&=  (m-1) \sum_{i=1}^m P_{\newinstance}(Y_{i}=1)  +  \sum_{1 \leq i \neq j \leq m} P_{\newinstance}(Y_{i}=1, Y_{j}=1) \, .
\end{align*}
Thus for any pair of rank $\sigma_1$ and $\sigma_2$, we have
\begin{align*}
  \mathbb{E}\left[ \ell_R(\sigma_2, \cdot) - \ell_R(\sigma_1, \cdot) \right | X=\newinstance] =   &\frac{1}{2}\mathbb{E}\left[ \ell_R(\sigma_2, \cdot) - \ell_R(\overline{\sigma}_2, \cdot)  \middle| X=\newinstance \right] + \frac{1}{2}\mathbb{E}\left[c(\cdot)  \middle| X=\newinstance \right] \\
    - &\frac{1}{2}\mathbb{E}\left[ \ell_R(\sigma_1, \cdot) - \ell_R(\overline{\sigma}_1, \cdot)  \middle| X=\newinstance \right] - \frac{1}{2}\mathbb{E}\left[c(\cdot)  \middle| X=\newinstance \right]  \\
    =&\frac{1}{2}\mathbb{E}\left[ \ell_R(\sigma_2, \cdot) - \ell_R(\overline{\sigma}_2, \cdot)  \middle| X=\newinstance \right] - \frac{1}{2}\mathbb{E}\left[ \ell_R(\sigma_1, \cdot) - \ell_R(\overline{\sigma}_1, \cdot)  \middle| X=\newinstance \right]
    \\
    =&\frac{1}{2}\sum_{i=1}^m (2\sigma^{-1}_2(i)-(m+1))P_{\newinstance}(Y_{i}=1) - \frac{1}{2}\sum_{i=1}^m (2\sigma^{-1}_1(i)-(m+1))P_{\newinstance}(Y_{i}=1)\\
    =&\frac{1}{2}\sum_{i=1}^m ((2\sigma^{-1}_2(i)-(m+1)) - (2\sigma^{-1}_1(i)-(m+1)) )P_{\newinstance}(Y_{i}=1) \\
    =&\frac{1}{2}\sum_{i=1}^m (2\sigma^{-1}_2(i) - 2\sigma^{-1}_1(i))P_{\newinstance}(Y_{i}=1) \\
    =& \sum_{i=1}^m (\sigma^{-1}_2(i) - \sigma^{-1}_1(i))P_{\newinstance}(Y_{i}=1) \\\
    =&\sum_{i\in \setindicesind_{\sigma_2\neq\sigma_1}} P_{\newinstance}(Y_{i}=1) (\sigma^{-1}_2(i) - \sigma^{-1}_1(i)) \,.
\end{align*} 
This implies 
\begin{align*}
\sigma_1 \sqsupset_{\lossrank, \credalind} \sigma_2 
 \iff	\inf_{P_{\newinstance} \in\credalind}\sum_{
		i\in\setindicesind^{\sigma_1\neq\sigma_2}} 
	P_{\newinstance}(Y_{i}=1) 
	(\sigma^{-1}_2(i) - \sigma^{-1}_1(i)) >0\, . 
\end{align*}
where $\setindicesind^{\sigma_2\neq\sigma_1}$ is defined as $\setindicesind^{\sigma_2\neq\sigma_1}=\left\{i \middle| \sigma^{-1}_2(i) \neq \sigma^{-1}_1(i) \right\}$.

We now show that, under the label independence assumption of Equation~\eqref{eq:independence} and the assumption on the credal set of Equation~\eqref{eq:credal_joint_independence}, we have
\begin{align}\label{eq:equivalent_claim}
\inf_{P_{\newinstance} \in\credalind}\sum_{
		i\in\setindicesind^{\sigma_1\neq\sigma_2}} 
	P_{\newinstance}(Y_{i}=1) 
	(\sigma^{-1}_2(i) - \sigma^{-1}_1(i)) >0 \nonumber
\iff 
&\sum_{
		i\in\setindicesind^{\sigma_1\neq\sigma_2}} 
	\inf_{P_{\newinstance}(Y_{i}=1) \in\credal_i}P_{\newinstance}(Y_{i}=1) 
	(\sigma^{-1}_2(i) - \sigma^{-1}_1(i)) >0\nonumber \\
\iff 
&\sum_{
		i\in\setindicesind^{\sigma_1\neq\sigma_2}} 
	P_{\newinstance}^*(Y_{i}=1) 
	(\sigma^{-1}_2(i) - \sigma^{-1}_1(i)) >0	\, ,
	\end{align}
where
\begin{align}\label{eq:min_marginal}
 	P_{\newinstance}^*(Y_{i}=1) =
 	\begin{cases}
 	 	\underline{P}_{\newinstance}(Y_{i}=1) \text{ if }  \sigma^{-1}_2(i) > \sigma^{-1}_1(i) \,,\\
 	 	\overline{P}_{\newinstance}(Y_{i}=1) \text{ otherwise. } \\
 	\end{cases}
\end{align}
Let us remind that under the assumption of Equation~\eqref{eq:independence} and \eqref{eq:credal_joint_independence}, we can show that for any $\left(P_{\newinstance}(Y_{1}=1), \ldots , P_{\newinstance}(Y_{m}=1) \right) \in \credal_1 \times \ldots \times \credal_m$, 
\begin{align*}
     \sum_{\vect{y} \in \mathcal{Y}}P_{\newinstance}(\vect{y}) =\sum_{\vect{y} \in \mathcal{Y}} \prod_{i=1}^m P_{\newinstance}(Y_{i}=y_i) =   1 \, ,
\end{align*}
or, in other words, 
\begin{align*}
P_{\newinstance}:=\left\{P_{\newinstance}(\vect{y}) = \prod_{i=1}^m P_{\newinstance}(Y_{i}=y_i) ~\middle|~ \forall \vect{y} \in \mathcal{Y}\right\}     
\end{align*}
is a possible distribution. The relation of Equation~\eqref{eq:min_marginal} simply means that for any $	P'_{\newinstance} \neq 	P^*_{\newinstance}$, we have that
\begin{align*}
    \left(
	P'_{\newinstance}(Y_{i}=1) - P_{\newinstance}^*(Y_{i}=1) \right)
	(\sigma^{-1}_2(i) - \sigma^{-1}_1(i)) \geq 0 \, , \forall i\in\setindicesind^{\sigma_1\neq\sigma_2}.
\end{align*}
Now, suppose that, contrary to the claim of Equation~\eqref{eq:equivalent_claim}, we necessarily have the following case:
\begin{align}\label{eq:counter_example}
    \exists &\left(P'_{\newinstance}(Y_{1}=1), \ldots , P'_{\newinstance}(Y_{m}=1) \right) \in \credal_1 \times \ldots \times \credal_m \, , \nonumber\\
    \text{ s.t. } & \mathbb{E}_{P'_{\newinstance}}\left[ \ell_R(\sigma_2, \cdot) - \ell_R(\sigma_1, \cdot) \right | X=\newinstance] < \mathbb{E}_{P^{*}_{\newinstance}}\left[ \ell_R(\sigma_2, \cdot) - \ell_R(\sigma_1, \cdot) \right | X=\newinstance]\, .
\end{align}
It is clear that Equation~\eqref{eq:counter_example} leads to a contradiction because it leads to 
\begin{align*}
& \mathbb{E}_{P'_{\newinstance}}\left[ \ell_R(\sigma_2, \cdot) - \ell_R(\sigma_1, \cdot) \right | X=\newinstance] < \mathbb{E}_{P^*_{\newinstance}}\left[ \ell_R(\sigma_2, \cdot) - \ell_R(\sigma_1, \cdot) \right | X=\newinstance]   \\ 
\iff 
&\sum_{
		i\in\setindicesind^{\sigma_1\neq\sigma_2}} 
	P'_{\newinstance}(Y_{i}=1) 
	(\sigma^{-1}_2(i) - \sigma^{-1}_1(i)) < \sum_{
		i\in\setindicesind^{\sigma_1\neq\sigma_2}} 
	P_{\newinstance}^*(Y_{i}=1) 
	(\sigma^{-1}_2(i) - \sigma^{-1}_1(i))\\	
\iff 
&\sum_{i\in\setindicesind^{\sigma_1\neq\sigma_2}} \left(
	P'_{\newinstance}(Y_{i}=1) - P_{\newinstance}^*(Y_{i}=1) \right)
	(\sigma^{-1}_2(i) - \sigma^{-1}_1(i)) < 0 \, .
\end{align*}
This completes the proof.
\end{proof}


\begin{proof}[\bf Proof of Proposition ~\ref{prop:imprecise_ranks}]
For any pair of labels $(\lambda_i,\lambda_j)$, denote by $\sigma_1 = \{\lambda_i \succ \lambda_j\}$ and $\sigma_2 = \{\lambda_j \succ \lambda_i\}$. Lemma \ref{lem:pairwise_difference} implies~that
\begin{align}
    \sigma_1 \sqsupset_{\lossrank, \credalind} \sigma_2 
&\iff 
\sum_{
		i\in\setindicesind^{\sigma_1\neq\sigma_2}} 
	P_{\newinstance}^*(Y_{i}=1) 
	(\sigma^{-1}_2(i) - \sigma^{-1}_1(i)) >0 	\nonumber\\
&\iff 
	P_{\newinstance}^*(Y_{i}=1) 
	(\sigma^{-1}_2(i) - \sigma^{-1}_1(i)) + 	P_{\newinstance}^*(Y_{j}=1) 
	(\sigma^{-1}_2(j) - \sigma^{-1}_1(j)) >0 \nonumber\\
&\iff 
	\underline{P}_{\newinstance}(Y_{i}=1) 
	(2- 1) + 	\overline{P}_{\newinstance}(Y_{j}=1) 
	(1 - 2) >0 \nonumber	\\
&\iff 
	\underline{P}_{\newinstance}(Y_{i}=1) 
	- 	\overline{P}_{\newinstance}(Y_{j}=1)  >0 \nonumber\\
&\iff 
	\underline{P}_{\newinstance}(Y_{i}=1) 
	>	\overline{P}_{\newinstance}(Y_{j}=1)  	\, .
\end{align}

We thus introduce the binary relation $R$ on the set $\mathcal{Y}$ such that
\begin{align}
\begin{cases}
 \lambda_i \succ_{R} \lambda_j\, \text{ if } \underline{P}_{\newinstance}(y_i=1)  >  \overline{P}_{\newinstance}(y_j=1) \, , i \neq j  \, , \\
 \lambda_i \succ_{R} \lambda_i \, , i \in [m] \, .
\end{cases}
\end{align}
It can be easily verified that $R$ is a  partial order as the reflexivity, transitivity and antisymmetry hold.  

This partial order then allows us to derive interval rank values \citep[Sec. $4.1$]{patil2004multiple} as we have that
\begin{align}
  \lambda_i \succ \lambda_j \Rightarrow \sigma(i) \leq \sigma(j) \, .  
\end{align}

Once the relation $\succ$ is determined, $\mathcal{Y}$ is a poset (partially ordered set) and the corresponding relation matrix, denoted by $\zeta$, is a $m \times m$ matrix defined as 
\begin{align*}
\zeta_{i,j} = 
\begin{cases}
1 \text{ if } \lambda_i \succ \lambda_j \\
0 \text{ otherwise. }
\end{cases}
\end{align*}
The results given by Theorems 1 and 2 in \citep[Sec. $4.1$]{patil2004multiple} imply that each label $\lambda_i$ can be associated to an imprecise rank $[\underline{\sigma}_i, \overline{\sigma}_i]$ such that 
\begin{align*}
\overline{\sigma}_i & = m+1 -  \sum_{j=1}^m\zeta_{i,j} \, ,\\
\underline{\sigma}_i &=  \sum_{j=1}^m\zeta_{j,i} \, ,
\end{align*}
where $\underline{\sigma}_i$ and $\overline{\sigma}_i$ are respectively the smallest and largest rank that can given to $\lambda_i$ by any $\hat{\sigma}^{P}_{\ell_R} \in \hat{\mathbb{R}}^E_{\ell_R,\credalind}$. Furthermore, for any $j \in \sigma(i)$, Theorem $1$ in \citep[Sec. $4.1$]{patil2004multiple} implies that there is $P \in \credal$ and $\sigma=\hat{\sigma}^{P}_{\ell_R}$ s.t. $\sigma^{-1}(i) = j$. 
\end{proof}


\begin{proof}[\bf Proof of Proposition~\ref{prop:equality_relation_IE}]
For any $\sigma \in \hat{\mathbb{R}}^E_{\ell_R,\credal}$, it is clear that \begin{align*}
    &\sigma^{-1}(i) \in [\underline{\sigma}_i, \overline{\sigma}_i]\, , \forall i \in [m] \, , \\
    &\sigma^{-1}(i) \neq \sigma^{-1}(j) \, , \forall i \neq j \, .
\end{align*}
This means that $\sigma \in \hat{\mathbb{R}}^{LE}_{\ell_R,\credalind}$. In other words, 
\begin{align*}
   \hat{\mathbb{R}}^{LE}_{\ell_R,\credalind} \supset   \hat{\mathbb{R}}^E_{\ell_R,\credalind} \,.
\end{align*}
The proof of Proposition~\ref{prop:equality_relation_IE} is completed by showing that   
\begin{align*}
   \hat{\mathbb{R}}^{LE}_{\ell_R,\credalind} \subset   \hat{\mathbb{R}}^E_{\ell_R,\credalind} \, .
\end{align*}
This is done by showing that for any $\sigma \in \hat{\mathbb{R}}^{LE}_{\ell_R,\credalind} $, $\exists P \in \credalind$ s.t.   
\begin{align*}
    P_{\sigma(1)} \geq \ldots \geq P_{\sigma(m)} \, .
\end{align*}
Such a $P$ can be found using Algorithm \ref{alg:ranking2distribution}. This algorithm simply finds precise positions of the labels in the Hasse diagram (cf. \citep[Sec. $4.1$]{patil2004multiple}) where ranks of the labels are given by the linear extension $\sigma$ of the finite poset $\mathcal{Y}$.  
\SetAlCapHSkip{0em}
\DecMargin{-1em}
\begin{algorithm}[!ht]
  \SetKw{KwBy}{by}
  \DontPrintSemicolon
  \caption{Find a $P \in \credal$}\label{alg:ranking2distribution}
  \LinesNumbered
  \SetKwFunction{main}{main}
  \SetKwFunction{updateProbs}{updateProbs}
  \SetKwFunction{genProbs}{genProbs}
  \KwData{$\sigma$, $m$, $\credal$}
  \KwResult{$P$}
    \SetKwProg{myproc}{Procedure}{}{end}
  \myproc{\updateProbs{$\credal$, $m$, $\sigma(i)$}}{
\For{$j \in [m]$}{
    		$\left[\underline{P}_{\newinstance}(y_j=1), \overline{P}_{\newinstance}(y_j=1)  \right] = \left[\underline{P}_{\newinstance}(y_j=1), \overline{P}_{\newinstance}(y_j=1)  \right] \setminus \left[ +\infty , \overline{P}_{\newinstance}(y_{\sigma(i)}=1)   \right)$\\
    	}
  }{}
  \SetKwProg{myproc}{Procedure}{}{end}
  \myproc{\genProbs{$\sigma$, $m$, $\credal$}}{
$P = \{ 0, \ldots, 0\}$ \;
\For{$i \in [m]$}{
    		$P[\sigma(i)] =  \overline{P}_{\newinstance}(y_{\sigma(i)} =1)$\\
    		\updateProbs{$\credal$, $m$, $\sigma(i)$}\\
    	}
  }{}
  \SetKwProg{myalg}{Algorithm}{}{end}
  \myalg{\main{}}{
   	\genProbs{$\sigma$, $m$, $\credal$}\\
    \KwRet $P$
  }
\end{algorithm}

Altogether, we have that $\hat{\mathbb{R}}^{LE}_{\ell_R,\credalind} =   \hat{\mathbb{R}}^E_{\ell_R,\credalind}$.
\end{proof}

\begin{proof}[\bf Proof of Proposition~\ref{prop:equality_relation}]
Note that \citet{troffaes07decision} showed that for any credal the set of solution given by E-admissibility is a subset of the one given by the Maximality, i.e.
\begin{align*}
   \hat{\mathbb{R}}^E_{\lossrank,\credalind} \subset  \hat{\mathbb{R}}^M_{\lossrank,\credalind} \, .
\end{align*}
Thus, the proof of Proposition~\ref{prop:equality_relation} is completed by showing that 
\begin{align*}
   \hat{\mathbb{R}}^E_{\lossrank,\credalind} \supset  \hat{\mathbb{R}}^M_{\lossrank,\credalind} \, .
\end{align*}
Since $\hat{\mathbb{R}}^E_{\lossrank,\credalind} = \hat{\mathbb{R}}^{LE}_{\lossrank,\credalind}$, it is reduced to show that for all $\sigma \in \hat{\mathbb{R}}^M_{\lossrank,\credalind}$ we have 
\begin{align}\label{eq:equality_relation_1}
  \sigma^{-1}(i) \in [\underline{\sigma}_i, \overline{\sigma}_i] \, , i \in [m] \, .  
\end{align}

Now, suppose that, contrary to the claim \eqref{eq:equality_relation_1},  $\exists \sigma \in \hat{\mathbb{R}}^M_{\lossrank,\credalind}$ and there is (at least) one index $i$ such that $\sigma^{-1}(i) \not\in [\underline{\sigma}_i, \overline{\sigma}_i]$. Then we necessarily have at least one of the following cases:
\begin{align}
	 \text{(i)}~\sigma^{-1}(i) < \underline{\sigma}_i
	 \qquad\text{or}\qquad
	 \text{(ii)}~\sigma^{-1}(i) > \overline{\sigma}_i
\end{align}
The proof is completed by showing that both (i) and (ii) lead to contradiction in what follows:
\begin{itemize}
    \item[(i)] if $\sigma^{-1}(i) < \underline{\sigma}_i$.\\\mbox{}\\
    We have, according to Proposition~\ref{prop:imprecise_ranks}, $\underline{\sigma}_i$ labels $\lambda_j$ verifying the first condition of Equation~\eqref{eq:zeta_matrix} 
$\zeta_{j,i}=1$ (i.e. $\underline{P}_{\newinstance}(Y_j=1) > \overline{P}_{\newinstance}(Y_i=1)$), and since by assumption $\sigma^{-1}(i) < \underline{\sigma}_i$, there is always an index $j$ such that 
    \begin{align*}
  \underline{P}_{\newinstance}(Y_j=1)  > \overline{P}_{\newinstance}(Y_i=1) \, \text{ and } \sigma^{-1}(i) < \sigma^{-1}(j).  
    \end{align*}
    This means that we get
    \begin{align}\label{eq:constraint_1}
        \inf_{P\in\credalind}P_{\newinstance}(Y_j=1)  - P_{\newinstance}(Y_i=1) > 0 \, \quad\text{and}\quad \lambda_i \succ_\sigma \lambda_j
    \end{align}
    
    Besides,  we can always find a ranking $\pi$ where only the ranks of labels $\lambda_i, \lambda_j$ are swapped, i.e. 
    \begin{align*}
		\pi^{-1}(k) = 
		\begin{cases}
		\sigma^{-1}(k)&\text{if}\quad k \not\in \{i, j \} ,\\
		\sigma^{-1}(i)&\text{if}\quad k =  j \, ,\\
		\sigma^{-1}(j)&\text{if}\quad k =  i \, . \\
		\end{cases}
	\end{align*}
	and therefore the set of disagreement indices between $\pi$ and $\sigma$ is $\setindices_{\pi\neq\sigma} = \{(j,i)\}$. By applying Proposition~\ref{prop:reduction} (i.e. the maximality criterion), we have that 
	\begin{align*}
		\pi \sqsupset_{\lossrank, \credalind} \sigma \iff &
			\inf_{P\in\credalind} \sum_{(i,j) 
				\in \setindices_{\pi\neq\sigma}}  
	P_{\newinstance}(Y_{i}=1) -  P_{\newinstance}(Y_{j}=1)>0\\
	\iff & \inf_{P\in\credalind}~~ 
		P_{\newinstance}(Y_{j}=1) - P_{\newinstance}(Y_{i}=1)>0
	\end{align*}
	By using Equation~\eqref{eq:constraint_1}, the last equation is verified, and hence, $\sigma$ is dominated by $\pi$ (i.e. $\pi \sqsupset_{\lossrank, \credalind} \sigma$). In other words, $\sigma \not \in \hat{\mathbb{R}}^M_{\lossrank,\credalind}$, which is a contradiction.
	\item[(ii)] if $\sigma^{-1}(i) > \overline{\sigma}_i$\\\mbox{}\\
	We have, according to Proposition~\ref{prop:imprecise_ranks}, $(m+1-\overline{\sigma}_i)$ labels $\lambda_j$ verifying the first condition of Equation~\eqref{eq:zeta_matrix} 
$\zeta_{i,j}=1$ (i.e. $\underline{P}_{\newinstance}(Y_i=1) > \overline{P}_{\newinstance}(Y_j=1)$), and since by assumption $\sigma^{-1}(i) > \overline{\sigma}_i$, there is always an index $j$ such that 
	\begin{align*}
     \underline{P}_{\newinstance}(Y_i=1) > \overline{P}_{\newinstance}(Y_j=1) \, \text{ and } \sigma^{-1}(i) > \sigma^{-1}(j).  
    \end{align*}
	 This means that we get
	  \begin{align}\label{eq:constraint_2}
        \inf_{P\in\credalind}P_{\newinstance}(Y_i=1)  - P_{\newinstance}(Y_j=1) > 0 \, \quad\text{and}\quad \lambda_j \succ_\sigma \lambda_i
    \end{align}
    Besides,  we can always find a ranking $\tau$ where only the ranks of labels $\lambda_i, \lambda_j$ are swapped, i.e. 
	 \begin{align*}
		\tau(k) = 
		\begin{cases}
		\sigma^{-1}(k) &\text{if}\quad k \not\in \{i\, , j \} \, ,\\
		\sigma^{-1}(i) &\text{if}\quad k =  j \, ,\\
		\sigma^{-1}(j) &\text{if}\quad k =  i \, . \\
		\end{cases}
	\end{align*}
	and therefore the set of disagreement indices between $\tau$ and $\sigma$ is $\setindices_{\tau\neq\sigma} = \{(i,j)\}$. By applying Proposition~\ref{prop:reduction} (i.e. the maximality criterion), we have that 
	\begin{align*}
		\pi \sqsupset_{\lossrank, \credalind} \sigma \iff &
			\inf_{P\in\credalind} \sum_{(i,j) 
				\in \setindices_{\pi\neq\sigma}}  
	P_{\newinstance}(Y_{i}=1) -  P_{\newinstance}(Y_{j}=1)>0\\
	\iff & \inf_{P\in\credalind}~~ 
		P_{\newinstance}(Y_{i}=1) - P_{\newinstance}(Y_{j}=1)>0
	\end{align*}
	By using Equation~\eqref{eq:constraint_2}, the last equation is verified, and hence, $\sigma$ is dominated by $\tau$ (i.e. $\tau \sqsupset_{\lossrank, \credalind} \sigma$). In other words, $\sigma \not \in \hat{\mathbb{R}}^M_{\lossrank,\credalind}$, which is again a contradiction.
\end{itemize}

\end{proof}

\end{document}

%% file: notations.tex
\newcommand{\condset}[2]{{ \left\{ #1 \middle| #2 \right\} }}

\newcommand{\indicator}[1]{{\mathbbm{1}_{(#1)}\;}}

\newcommand{\credal}{{\mathcal{P}}}

\newcommand{\expe}{{\mathbb{E}}}
\newcommand{\lexpe}{{\underline{\expe}}}
\newcommand{\uexpe}{{\overline{\expe}}}

\newcommand{\lpr}{{\underline{P}}}
\newcommand{\upr}{{\overline{P}}}

\newcommand{\reals}{\mathbb{R}}

\newcommand{\insta}{{\mathbf{x}}}

%% file: eg_improb_loss_tree.tex
\newcommand{\exptwoeg}[3]{\underline{\mathbb{E}}_{Y_{\{#1\}}|\newinstance}\left[ \ell_H(\cdot, \overline{a}_\mathcal{I}) \middle| Y_{\{#2\}}\!=\!#3\right]}
\newcommand{\expeg}[1]{\underline{\mathbb{E}}_{Y_{\{#1\}}|\newinstance}\left[ \cdot \right]}
\newcommand{\condlowprob}{Y_{m} | Y_{\mathcal{I}_{\setn{m-1}}}}

\tikzset{
  treenode/.style = {shape=rectangle, rounded corners,
                     draw, align=center,
                     top color=white, bottom color=blue!20},
  root/.style     = {treenode, font=\Large, bottom color=red!30},
  env/.style      = {treenode, font=\ttfamily\footnotesize},
  dummy/.style    = {circle,draw},
  startstop/.style = {rectangle, draw, semithick, rounded corners=3mm,
                minimum width=#1, minimum height=1cm},
}
\tikzstyle{level 1}=[level distance=2cm, sibling distance=2.8cm]
\tikzstyle{level 2}=[level distance=2.5cm, sibling distance=1.4cm]
\tikzstyle{level 3}=[level distance=2.7cm, sibling distance=0.9cm]
\tikzstyle{end} = [inner sep=0pt]
\begin{tikzpicture}[
	grow=right, auto,
	sloped,
	edge from parent/.style = {draw, -latex, font=\tiny\sffamily, text=blue},
	every node/.style  = {font=\scriptsize}]
\node[red] at (2.05, 3.15) {$\bf Y_1$};
\draw [dashed, red] (2.00, 3.0) -- (2.00,-2.92);
\node[red] at (4.55, 3.15) {$\bf Y_2$};
\draw [dashed, red] (4.50, 3.0) -- (4.50,-2.92);
\node[red] at (7.25, 3.15) {$\bf Y_3$};
\draw [dashed, red] (7.20, 3.0) -- (7.20,-2.92);
\node[root] {}
    child {
        node[env] {$1$}        
            child {
                node[env] {$1$}
                child {
	                	node[env]{$1$}
	                	edge from parent
                		node[below, inner sep=1pt] (111) {$[0.07, \mathbf{0.37}]$}
                }
                child {
                		node[env]{$0$}
	                	edge from parent
                		node[above, inner sep=1pt] (110) {$[\mathbf{0.63}, 0.93]$}
                }
                edge from parent
                node[below, inner sep=1pt] (11) {$[0.54, \mathbf{0.84}]$}
            }
            child {
                node[env] {$0$}
                child {
	                	node[env]{$1$}
	                	edge from parent
                		node[below, inner sep=1pt] (101) {$[0.26, \mathbf{0.56}]$}
                }
                child {
                		node[env]{$0$}
	                	edge from parent
                		node[above, inner sep=1pt] (100) {$[\mathbf{0.44}, 0.74]$}
                }
                edge from parent
                node[above, inner sep=1pt] (10) {$[\mathbf{0.16}, 0.46]$}
            }
            edge from parent 
            node[below, inner sep=1pt] (1) {$[\mathbf{0.04}, 0.33]$} 
    }
    child {
        node[env] {$0$}       
        child {
                node[env] {$1$}
                child {
	                	node[env]{$1$}
	                	edge from parent
                		node[below, inner sep=1pt] (011) {$[0.33, \mathbf{0.63}]$}
                }
                child {
                		node[env]{$0$}
	                	edge from parent
                		node[above, inner sep=1pt] (010) {$[\mathbf{0.37}, 0.67]$}
                }
                edge from parent
                node[below] (01) {$[0.36, \mathbf{0.66}]$}
            }
        child {
                node[env] {$0$}
                child {
	                	node[env]{$1$}
	                	edge from parent
                		node[below, inner sep=1pt] (001) {$[0.63, \mathbf{0.93}]$}
                }
                child {
                		node[env]{$0$}
	                	edge from parent
                		node[above, inner sep=1pt] (000) {$[\mathbf{0.07}, 0.37]$}
                }
                edge from parent
                node[above, inner sep=1pt] (00)  {$[\mathbf{0.34}, 0.64]$}
            }
        edge from parent         
        node[above, inner sep=1pt] (0) {$[0.67, \mathbf{0.96}]$}
    };

\node[red] at (8.3, 3.15) {$\sigma' \sqsupset_{\loss_R, \credal} \sigma''$};
\node[blue, ] at (8.3, 2.55) {\footnotesize $0 - 0 = ~~0$};
\node[blue] at (8.3, 1.65) {\footnotesize$0 - 1 = -1$};
\node[blue] at (8.3, 1.15) {\footnotesize$0 - 1 = -1$};
\node[blue] at (8.3, 0.25) {\footnotesize$0 - 2 = -2$};
\node[blue] at (8.3, -0.25) {\footnotesize$2 - 0 = +2$};
\node[blue] at (8.3, -1.15) {\footnotesize$2 - 1 = +1$};
\node[blue] at (8.3, -1.65) {\footnotesize$2 - 1 = +1$};
\node[blue] at (8.3, -2.65) {\footnotesize$2 - 2 = ~~0$};

\draw[black, transform canvas={xshift=-22pt}, bend right,-] (001) to node [auto, yshift=-40pt, xshift=0pt] {\rotatebox{270}{$\lexpe\!=\!-0.93$}} (000);

\draw[black, transform canvas={xshift=-22pt}, bend right,-]  (011) to node [auto, yshift=-40pt, xshift=0pt] {\rotatebox{270}{$\lexpe\!=\!-1.63$}} (010);

\draw[black, transform canvas={xshift=-22pt}, bend right,-]  (101) to node [auto, yshift=-35pt, xshift=0pt] {\rotatebox{270}{$\lexpe\!=\!1.44$}} (100);

\draw[black, transform canvas={xshift=-22pt}, bend right,-]  (111) to node [auto, yshift=-35pt, xshift=0pt] {\rotatebox{270}{$\lexpe\!=\!0.63$}} (110);

\draw[black, transform canvas={xshift=-22pt}, bend right,-]  (11) to node [auto, yshift=-29pt, xshift=0pt] {\rotatebox{270}{$\lexpe\!=\!0.76$}} (10);

\draw[black, transform canvas={xshift=-22pt}, bend right,-]  (01) to node [auto, yshift=-2pt, xshift=0pt] {\rotatebox{88}{$\lexpe\!=\!-1.39$}} (00);

\draw[black, transform canvas={xshift=-20pt}, bend right,-]  (1) to node [auto, yshift=-100pt, xshift=0pt] {\rotatebox{-90}{$\lexpe\!=\!\mathbf{0.96} \cdot -1.39 + \mathbf{0.04} \cdot 0.76 < 0$}} (0);

\node [startstop=5mm, red] at (13, 1.5) {
	$\begin{aligned}
	 \lexpe_{Y_m|}[\cdot|Y_{\mathcal{I}_{\setn{m-1}}}] &= \min\left\{\begin{array}{@{}l}
        c_l^m*\underline{P}(\condlowprob) + c_r^m*\overline{P}(\condlowprob),\!\!\!\!\!\!\\
        c_l^m*\underline{P}(\condlowprob) + c_r^m*\overline{P}(\condlowprob) 
        \end{array}\!\!\!\right\}\\
        c_l^m &: \text{cost from the left node of label $Y_m$}\\
        c_r^m &: \text{cost from the right node of label $Y_m$}
	\end{aligned}$
};
\draw[-{Triangle[width=18pt, length=8pt]}, opacity=0.5, rotate=270, line width=10pt, red] (-0.6, 13) -- (0.7, 13);
\node [startstop=5mm, blue] (aa) at (13, -1.25) {
	$
	 \lexpe_{Y_3}[\lossrank(\cdot,?)] = \min\left\{\begin{array}{@{}l}
        0*0.07+ (-1)*0.93,\!\!\!\!\!\!\\
        0*0.63 + (-1)*0.37
        \end{array}\!\!\!\right\} = -0.93
    $
};
\draw [->, dashed, opacity=0.7] (6.3, 2.05) -- (aa);

\end{tikzpicture}

%% file: mlc_ranking_credal.bbl
\begin{thebibliography}{26}
\expandafter\ifx\csname natexlab\endcsname\relax\def\natexlab#1{#1}\fi
\providecommand{\url}[1]{\texttt{#1}}
\providecommand{\href}[2]{#2}
\providecommand{\path}[1]{#1}
\providecommand{\DOIprefix}{doi:}
\providecommand{\ArXivprefix}{arXiv:}
\providecommand{\URLprefix}{URL: }
\providecommand{\Pubmedprefix}{pmid:}
\providecommand{\doi}[1]{\href{http://dx.doi.org/#1}{\path{#1}}}
\providecommand{\Pubmed}[1]{\href{pmid:#1}{\path{#1}}}
\providecommand{\bibinfo}[2]{#2}
\ifx\xfnm\relax \def\xfnm[#1]{\unskip,\space#1}\fi
\bibitem[{Hayes and Weinstein(1990)}]{hayes1990construe}
\bibinfo{author}{P.~J. Hayes}, \bibinfo{author}{S.~P. Weinstein},
\newblock \bibinfo{title}{Construe/tis: A system for content-based indexing of
  a database of news stories},
\newblock in: \bibinfo{booktitle}{Proceedings of The Second Conference on
  Innovative Applications of Artificial Intelligence (IAAI)},
  \bibinfo{organization}{AAAI Press}, \bibinfo{year}{1990}, pp.
  \bibinfo{pages}{49--64}.
\bibitem[{Lewis(1992)}]{lewis1992evaluation}
\bibinfo{author}{D.~D. Lewis},
\newblock \bibinfo{title}{An evaluation of phrasal and clustered
  representations on a text categorization task},
\newblock in: \bibinfo{booktitle}{Proceedings of the 15th annual International
  ACM SIGIR Conference on Research and Development in Information Retrieval
  (SIGIR)}, \bibinfo{organization}{ACM}, \bibinfo{year}{1992}, pp.
  \bibinfo{pages}{37--50}.
\bibitem[{Trohdis(2008)}]{trohidis2008multi}
\bibinfo{author}{K.~Trohdis},
\newblock \bibinfo{title}{Multi-label classification of music into emotions},
\newblock in: \bibinfo{booktitle}{Proceedings of the 9th International
  Conference on Music Information Retrieval (ISMIR)}, \bibinfo{year}{2008}, pp.
  \bibinfo{pages}{325--330}.
\bibitem[{Boutell et~al.(2004)Boutell, Luo, Shen, and
  Brown}]{boutell2004learning}
\bibinfo{author}{M.~R. Boutell}, \bibinfo{author}{J.~Luo},
  \bibinfo{author}{X.~Shen}, \bibinfo{author}{C.~M. Brown},
\newblock \bibinfo{title}{Learning multi-label scene classification},
\newblock \bibinfo{journal}{Pattern recognition} \bibinfo{volume}{37}
  (\bibinfo{year}{2004}) \bibinfo{pages}{1757--1771}.
\bibitem[{Elisseeff and Weston(2001)}]{elisseeff2002kernel}
\bibinfo{author}{A.~Elisseeff}, \bibinfo{author}{J.~Weston},
\newblock \bibinfo{title}{A kernel method for multi-labelled classification},
\newblock in: \bibinfo{booktitle}{Proceedings of the 14th International
  Conference on Neural Information Processing Systems (NIPS)},
  \bibinfo{organization}{MIT Press}, \bibinfo{year}{2001}, pp.
  \bibinfo{pages}{681--687}.
\bibitem[{Zhang and Zhou(2013)}]{zhang2013review}
\bibinfo{author}{M.-L. Zhang}, \bibinfo{author}{Z.-H. Zhou},
\newblock \bibinfo{title}{A review on multi-label learning algorithms},
\newblock \bibinfo{journal}{IEEE transactions on knowledge and data
  engineering} \bibinfo{volume}{26} (\bibinfo{year}{2013})
  \bibinfo{pages}{1819--1837}.
\bibitem[{Tsoumakas et~al.(2010)Tsoumakas, Katakis, and
  Vlahavas}]{tsoumakas2010random}
\bibinfo{author}{G.~Tsoumakas}, \bibinfo{author}{I.~Katakis},
  \bibinfo{author}{I.~Vlahavas},
\newblock \bibinfo{title}{Random k-labelsets for multilabel classification},
\newblock \bibinfo{journal}{IEEE Transactions on Knowledge and Data
  Engineering} \bibinfo{volume}{23} (\bibinfo{year}{2010})
  \bibinfo{pages}{1079--1089}.
\bibitem[{Dembczynski et~al.(2012)Dembczynski, Kotlowski, and
  H{\"u}llermeier}]{dembczynski2012consistent}
\bibinfo{author}{K.~Dembczynski}, \bibinfo{author}{W.~Kotlowski},
  \bibinfo{author}{E.~H{\"u}llermeier},
\newblock \bibinfo{title}{Consistent multilabel ranking through univariate
  losses},
\newblock in: \bibinfo{booktitle}{Proceedings of the 29th International
  Conference on Machine Learning (ICML)}, \bibinfo{year}{2012}, pp.
  \bibinfo{pages}{1319--1326}.
\bibitem[{Jung and Tewari(2018)}]{jung2018online}
\bibinfo{author}{Y.~H. Jung}, \bibinfo{author}{A.~Tewari},
\newblock \bibinfo{title}{Online boosting algorithms for multi-label ranking},
\newblock in: \bibinfo{booktitle}{Proceedings of the 21st International
  Conference on Artificial Intelligence and Statistics (AISTATS)},
  \bibinfo{year}{2018}, pp. \bibinfo{pages}{279--287}.
\bibitem[{Kanehira and Harada(2016)}]{kanehira2016multi}
\bibinfo{author}{A.~Kanehira}, \bibinfo{author}{T.~Harada},
\newblock \bibinfo{title}{Multi-label ranking from positive and unlabeled
  data},
\newblock in: \bibinfo{booktitle}{Proceedings of the IEEE Conference on
  Computer Vision and Pattern Recognition (CVPR)}, \bibinfo{year}{2016}, pp.
  \bibinfo{pages}{5138--5146}.
\bibitem[{Carranza~Alarcón and Destercke(2022)}]{yonseb2020hammingmlc}
\bibinfo{author}{Y.~C. Carranza~Alarcón}, \bibinfo{author}{S.~Destercke},
\newblock \bibinfo{title}{Skeptical binary inferences in multi-label problems
  with sets of probabilities}  (\bibinfo{year}{2022}). \URLprefix
  \url{https://arxiv.org/abs/2205.00662}.
  \DOIprefix\doi{10.48550/ARXIV.2205.00662}.
\bibitem[{Nguyen and H{\"u}llermeier(2021)}]{nguyen2021multilabel}
\bibinfo{author}{V.-L. Nguyen}, \bibinfo{author}{E.~H{\"u}llermeier},
\newblock \bibinfo{title}{Multilabel classification with partial abstention:
  Bayes-optimal prediction under label independence},
\newblock \bibinfo{journal}{Journal of Artificial Intelligence Research}
  \bibinfo{volume}{72} (\bibinfo{year}{2021}) \bibinfo{pages}{613--665}.
\bibitem[{Pillai et~al.(2013)Pillai, Fumera, and Roli}]{pillai2013multi}
\bibinfo{author}{I.~Pillai}, \bibinfo{author}{G.~Fumera},
  \bibinfo{author}{F.~Roli},
\newblock \bibinfo{title}{Multi-label classification with a reject option},
\newblock \bibinfo{journal}{Pattern Recognition} \bibinfo{volume}{46}
  (\bibinfo{year}{2013}) \bibinfo{pages}{2256--2266}.
\bibitem[{Levi(1980)}]{levi1980a}
\bibinfo{author}{I.~Levi}, \bibinfo{title}{The Enterprise of Knowledge},
  \bibinfo{publisher}{MIT Press}, \bibinfo{address}{London},
  \bibinfo{year}{1980}.
\bibitem[{Troffaes(2007)}]{troffaes07decision}
\bibinfo{author}{M.~Troffaes},
\newblock \bibinfo{title}{Decision making under uncertainty using imprecise
  probabilities},
\newblock \bibinfo{journal}{Int. J. of Approximate Reasoning}
  \bibinfo{volume}{45} (\bibinfo{year}{2007}) \bibinfo{pages}{17--29}.
\bibitem[{Read et~al.(2021)Read, Pfahringer, Holmes, and Frank}]{Read_2021}
\bibinfo{author}{J.~Read}, \bibinfo{author}{B.~Pfahringer},
  \bibinfo{author}{G.~Holmes}, \bibinfo{author}{E.~Frank},
\newblock \bibinfo{title}{Classifier chains: A review and perspectives},
\newblock \bibinfo{journal}{Journal of Artificial Intelligence Research}
  \bibinfo{volume}{70} (\bibinfo{year}{2021}) \bibinfo{pages}{683--718}.
  \URLprefix \url{https://doi.org/10.1613%2Fjair.1.12376}.
  \DOIprefix\doi{10.1613/jair.1.12376}.
\bibitem[{Dembczy{\'n}ski et~al.(2012)Dembczy{\'n}ski, Waegeman, Cheng, and
  H{\"u}llermeier}]{dembczynski2012label}
\bibinfo{author}{K.~Dembczy{\'n}ski}, \bibinfo{author}{W.~Waegeman},
  \bibinfo{author}{W.~Cheng}, \bibinfo{author}{E.~H{\"u}llermeier},
\newblock \bibinfo{title}{On label dependence and loss minimization in
  multi-label classification},
\newblock \bibinfo{journal}{Machine Learning} \bibinfo{volume}{88}
  (\bibinfo{year}{2012}) \bibinfo{pages}{5--45}.
\bibitem[{Friedman et~al.(2001)Friedman, Hastie, and
  Tibshirani}]{friedman2001elements}
\bibinfo{author}{J.~Friedman}, \bibinfo{author}{T.~Hastie},
  \bibinfo{author}{R.~Tibshirani}, \bibinfo{title}{The elements of statistical
  learning}, \bibinfo{publisher}{Springer New York Inc.}, \bibinfo{year}{2001}.
\bibitem[{Augustin et~al.(2014)Augustin, Coolen, de~Cooman, and
  Troffaes}]{augustin2014introduction}
\bibinfo{author}{T.~Augustin}, \bibinfo{author}{F.~P. Coolen},
  \bibinfo{author}{G.~de~Cooman}, \bibinfo{author}{M.~C. Troffaes},
  \bibinfo{title}{Introduction to imprecise probabilities},
  \bibinfo{publisher}{John Wiley \& Sons}, \bibinfo{year}{2014}.
\bibitem[{Montes et~al.(2020)Montes, Miranda, and
  Destercke}]{montes2020unifying}
\bibinfo{author}{I.~Montes}, \bibinfo{author}{E.~Miranda},
  \bibinfo{author}{S.~Destercke},
\newblock \bibinfo{title}{Unifying neighbourhood and distortion models: Part
  i-new results on old models},
\newblock \bibinfo{journal}{International Journal of General Systems}
  (\bibinfo{year}{2020}).
\bibitem[{Rahimian and Mehrotra(2019)}]{rahimian2019distributionally}
\bibinfo{author}{H.~Rahimian}, \bibinfo{author}{S.~Mehrotra},
\newblock \bibinfo{title}{Distributionally robust optimization: A review},
\newblock \bibinfo{journal}{arXiv preprint arXiv:1908.05659}
  (\bibinfo{year}{2019}).
\bibitem[{Destercke(2015)}]{destercke2015multilabel}
\bibinfo{author}{S.~Destercke},
\newblock \bibinfo{title}{Multilabel predictions with sets of probabilities:
  the hamming and ranking loss cases},
\newblock \bibinfo{journal}{Pattern Recognition} \bibinfo{volume}{48}
  (\bibinfo{year}{2015}) \bibinfo{pages}{3757--3765}.
\bibitem[{Hermans et~al.(2009)Hermans, Quaeghebeur
  et~al.}]{hermans2009imprecise}
\bibinfo{author}{F.~Hermans}, \bibinfo{author}{E.~Quaeghebeur}, et~al.,
\newblock \bibinfo{title}{Imprecise markov chains and their limit behavior},
\newblock \bibinfo{journal}{Probability in the Engineering and Informational
  Sciences} \bibinfo{volume}{23} (\bibinfo{year}{2009})
  \bibinfo{pages}{597--635}.
\bibitem[{Alarc{\'o}n and Destercke(2021)}]{alarcon2021distributionally}
\bibinfo{author}{Y.~C.~C. Alarc{\'o}n}, \bibinfo{author}{S.~Destercke},
\newblock \bibinfo{title}{Distributionally robust, skeptical binary inferences
  in multi-label problems},
\newblock in: \bibinfo{booktitle}{International Symposium on Imprecise
  Probability: Theories and Applications}, \bibinfo{organization}{PMLR},
  \bibinfo{year}{2021}, pp. \bibinfo{pages}{51--60}.
\bibitem[{Gent et~al.(2008)Gent, Miguel, and Nightingale}]{gent2008generalised}
\bibinfo{author}{I.~P. Gent}, \bibinfo{author}{I.~Miguel},
  \bibinfo{author}{P.~Nightingale},
\newblock \bibinfo{title}{Generalised arc consistency for the alldifferent
  constraint: An empirical survey},
\newblock \bibinfo{journal}{Artificial Intelligence} \bibinfo{volume}{172}
  (\bibinfo{year}{2008}) \bibinfo{pages}{1973--2000}.
\bibitem[{Patil and Taillie(2004)}]{patil2004multiple}
\bibinfo{author}{G.~Patil}, \bibinfo{author}{C.~Taillie},
\newblock \bibinfo{title}{Multiple indicators, partially ordered sets, and
  linear extensions: Multi-criterion ranking and prioritization},
\newblock \bibinfo{journal}{Environmental and ecological statistics}
  \bibinfo{volume}{11} (\bibinfo{year}{2004}) \bibinfo{pages}{199--228}.

\end{thebibliography}
